\documentclass[pdflatex,sn-mathphys,Numbered]{sn-jnl}

\usepackage{graphicx}%
\usepackage{multirow}%
\usepackage{amsmath,amssymb,amsfonts}%
\usepackage{amsthm}%
\usepackage{mathrsfs}%
\usepackage[title]{appendix}%
\usepackage{xcolor}%
\usepackage{textcomp}%
\usepackage{manyfoot}%
\usepackage{booktabs}%
\usepackage{algorithm}%
\usepackage{algorithmicx}%
\usepackage{algpseudocode}%
\usepackage{listings}%
\usepackage{longtable}
\begin{document}

	\title[Article Title]{Deep Learning for Cervical Cytology Screening: Methods, Algorithms, Challenges and Prospects}


	\author[1,2]{\fnm{Yangbo} \sur{Jiang}}\email{jiangyangbo@zju.edu.cn}


	\author*[1,2]{\fnm{Nenggan} \sur{Zheng}}\email{zng@cs.zju.edu.cn}

	\affil*[1]{\orgdiv{Qiushi Academy for Advanced Studies}, \orgname{Zhejiang University}, \orgaddress{\street{38 Zheda Road}, \city{Hangzhou}, \postcode{310012}, \country{China}}}

	\affil[2]{\orgdiv{College of Computer Science and Technology}, \orgname{Zhejiang University}, \orgaddress{\street{38 Zheda Road}, \city{Hangzhou}, \postcode{310012}, \country{China}}}




\abstract{Cervical cancer remains a leading cause of cancer death among women, with more than 90\% of deaths occurring in low- and middle-income countries, where cytology screening is constrained by a shortage of trained cytopathologists. Deep learning has made automated cytology screening increasingly feasible, but the literature is fragmented across cell, patch and whole-slide-image scales, task formulations and supervision regimes, hindering comparison and clinical translation. This Review synthesizes the deep learning literature for cervical cytology screening, organizing the field along three axes---data scale, functional task and supervision---and distinguishing two clinical roadmaps: whole-slide-image confidence aggregation and cell-level detection with Bethesda classification. We examine 16 publicly available or publicly documented cytology datasets alongside the CAISHI histopathology dataset, compare quantitative performance while exposing incomparable metrics and data leakage, and assess commercial systems, regulatory status, HPV-era screening pathways and low-resource deployment. Single-cell accuracy on Herlev and SIPaKMeD has plateaued, but this reflects benchmark saturation rather than clinical readiness; prospective and randomized evidence is emerging but remains limited. We consolidate the evidence into five challenge areas---data and benchmarks, architecture and label-efficient learning, trustworthy AI, clinical translation and health economics, and evaluation and reproducibility---and provide an evidence-to-agenda map and a deployment decision matrix. Progress now depends on multi-center benchmarks, externally validated label-efficient methods, calibrated and fair predictions, and pragmatic clinical trials rather than further gains on saturated single-cell benchmarks.}


	\maketitle

	\section{Introduction} \label{sec1}



Cervical cancer caused approximately 660,000 new cases and 350,000 deaths in 2022, ranking fourth among cancers in women globally \cite{bray2024global}. In many low- and middle-income countries, cervical cancer is among the leading causes of cancer death among women, driven primarily by inadequate access to early screening and treatment. Sub-Saharan Africa, Central America, and Southeast Asia bear the highest incidence and mortality rates, with more than 90\% of cervical cancer deaths concentrated in low- and middle-income countries \cite{bray2024global}. Early screening detects precancerous lesions and prevents up to 80\% of cervical cancer cases, making cervical cytology screening one of the most effective cancer prevention interventions available \cite{Who2021human}. Automated analysis of cervical cytology images using computer vision and deep learning has emerged as a critical tool for addressing the global cytopathologist shortage, particularly in low-resource settings where the disease burden is highest.


Several surveys have examined automated cervical cytology screening from various perspectives. Earlier reviews covered traditional image processing and classifier-based methods \cite{Duanggate2008,plissiti2013review,William2018c,Conceicao2019,Sarwar2020,Chitra2021}, while more recent surveys addressed deep learning-based approaches \cite{VictoriaMatias2021,jiang2023systematic}. Fang et al. systematically reviewed deep learning methods for cervical cell image analysis, covering CNNs, transformers, datasets, and metrics \cite{fang2024systematic}. A 2026 systematic review of 77 studies reported a mean classification accuracy of 87.76\%, with only a subset of hybrid and vision-transformer models exceeding 92\% \cite{valles2026application}. A separate 2025 meta-analysis of 77 studies reported that AI-assisted Pap smear interpretation achieved 94\% accuracy and 95\% sensitivity, with substantial heterogeneity across study designs \cite{liu2025performance}.

However, prior surveys have not organized methods from a data-level perspective that reflects both the technical evolution (from handcrafted features to end-to-end learning) and the clinical workflow (from individual cells to gigapixel whole-slide images). The rapid emergence of foundation models, self-supervised learning paradigms, graph neural networks, and state-space model architectures during 2024--2026 demands a fresh synthesis. This survey addresses these gaps through a three-axis taxonomy---data scale, functional task, and supervision---and two clinical roadmaps that connect methods to deployment. The survey focuses on deep learning; classical image processing and classifier-based methods are retained only as background and are not the subject of the taxonomy.

The main contributions of this study are five-fold:
\begin{itemize}
	\item We propose a three-axis taxonomy (data scale $\times$ functional task $\times$ supervision) and distinguish two clinical roadmaps---WSI confidence aggregation and cell detection---that connect methods to deployment.
	\item We provide a structured review of 16 publicly available or publicly documented cytology datasets plus the CAISHI histopathology dataset, with a comparison table covering scale, cytology type, annotation, annotators, tasks, and limitations.
	\item We organize methods by task at cell, patch, and WSI scales, add a cross-scale synthesis of weak/semi-supervised, self-supervised, few-shot, foundation-model, and domain-adaptation paradigms, and compare quantitative performance while flagging incomparable metrics.
	\item We analyze clinical translation, including commercial systems, FDA/NMPA/CE status, prospective and randomized evidence, HPV-era screening pathways, and low-resource and health-economic constraints.
	\item We consolidate the evidence into five challenge areas, provide an evidence-to-agenda map (Table~\ref{tbl:tlb13}), and conclude with a deployment decision matrix and a prioritized research agenda.
\end{itemize}


This paper is structured as follows. Section~\ref{sec:clinical} defines the clinical background, screening pathways, and the two technical roadmaps. Section~\ref{sec:review_methodology} describes the literature search and screening methodology. Section~\ref{sec:framework} presents the three-axis taxonomy and the evaluation and reporting framework. Section~\ref{sec:datasets} surveys datasets and benchmarks. Sections~\ref{sec:cell}--\ref{sec:wsi} review cell-, patch-, and WSI-level methods, with WSI methods organized by classification/aggregation and detection/localization tasks. Section~\ref{sec:paradigms} synthesizes learning paradigms across scales. Section~\ref{sec:translation} examines clinical translation, regulation, and health economics. Section~\ref{sec:trust} covers trustworthiness, fairness, and human--AI interaction. Section~\ref{sec:challenges} consolidates five challenge areas and research directions, and Section~\ref{sec:conclusion} concludes with a decision matrix and priority agenda.
\section{Clinical Background and Screening Pathways}
\label{sec:clinical}

Automated cervical cytology is embedded in a clinical workflow that starts with sample collection and ends with a Bethesda-based diagnosis. This section defines the problem, summarizes the screening pathways that determine the operating point of AI systems, and introduces the two technical roadmaps used throughout the survey.

	\subsection{Problem Definition and Clinical Workflow}

\begin{figure*}
		\centering
		\includegraphics[width=\linewidth]{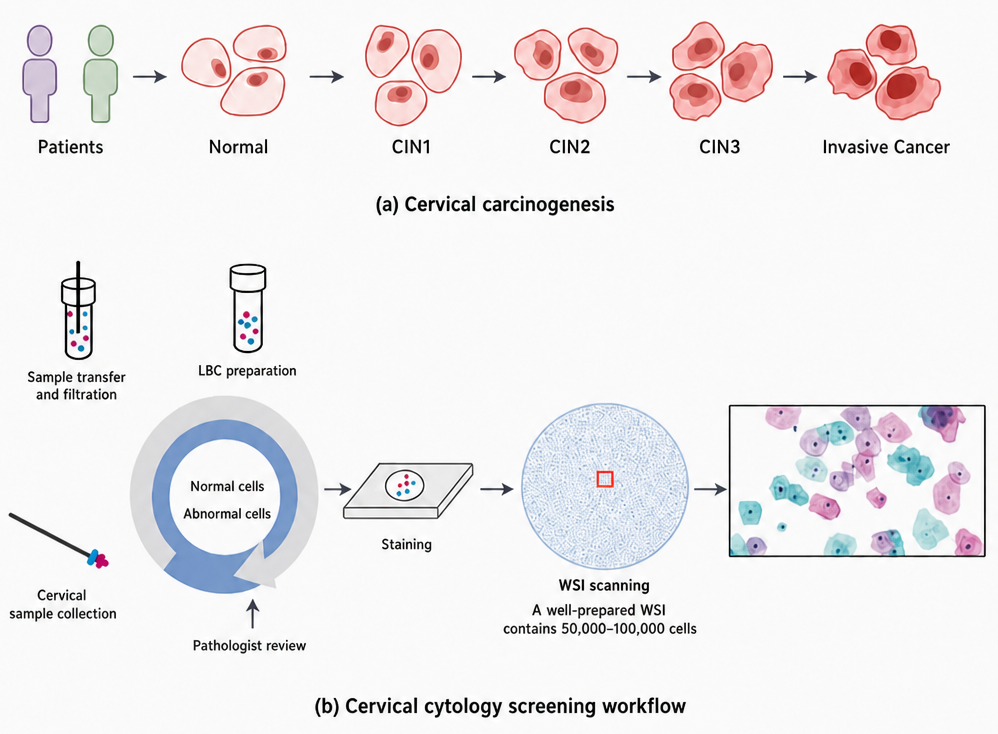}
		\caption{Illustration of the cervical cytology screening pipeline using liquid-based cytology (LBC). (a) Sample collection, LBC preparation, staining, and whole-slide image (WSI) scanning. (b) Review of the WSI and cellular details by a pathologist, followed by diagnosis according to the Bethesda System. Cell and WSI illustrations are schematic and do not represent acquired microscopy images.}
		\label{fig:fig1}
	\end{figure*}

Cervical cytology screening detects precancerous lesions through microscopic examination of exfoliated cervical epithelial cells. Cervical cancer is caused by persistent infection with high-risk subtypes of Human Papillomavirus (HPV), a virus comprising over 200 subtypes. Most HPV infections are transient and cleared by the immune system without intervention. However, high-risk subtypes---notably HPV-16 and HPV-18---induce abnormal cellular changes that can progress to malignancy. According to the WHO classification, cervical epithelial tumors fall into three categories: (1) squamous cell carcinoma (SCC), (2) adenocarcinoma (AC), and (3) other epithelial tumors. SCC accounts for 70\%--80\% of cases and is predominantly linked to HPV-16 infection \cite{marth2017cervical}. Adenocarcinoma constitutes approximately 15\% of cases and is associated with HPV-18 \cite{green2003risk}. The overall screening pipeline and the points at which automated methods intervene are illustrated in Fig.~\ref{fig:fig1}.

The core objective of automated cervical cytology screening is detecting abnormal cells on digitized slides to identify cervical cancer or precancerous lesions. This requires analyzing cell morphology, texture, and structural organization across multiple spatial scales.

\begin{figure*}
		\centering
		\includegraphics[width=\linewidth]{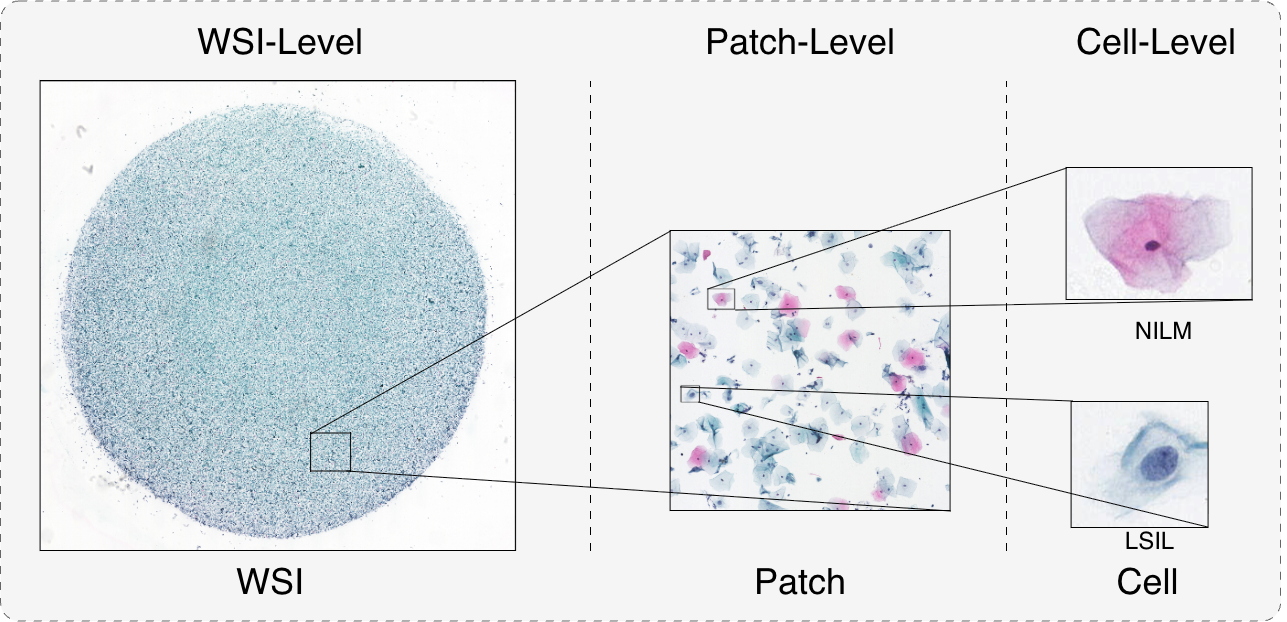}
		\caption{Illustration of the cervical cytology screening at three hierarchical levels.}
		\label{fig:fig2}
	\end{figure*}

From a data perspective, the analysis operates at three hierarchical levels, which is illustrated in Fig.~\ref{fig:fig2}. \textbf{Cell-level:} Individual cervical cells are classified according to the Bethesda System (TBS), which defines the following abnormal categories---Atypical Squamous Cells of Undetermined Significance (ASC-US), Atypical Squamous Cells cannot exclude HSIL (ASC-H), Low-grade Squamous Intraepithelial Lesion (LSIL), High-grade Squamous Intraepithelial Lesion (HSIL), Squamous Cell Carcinoma (SCC), Adenocarcinoma, Atypical Glandular Cells (AGC), and Adenocarcinoma In Situ (AIS). \textbf{Patch-level:} Image patches (typically 256$\times$256 to 1024$\times$1024 pixels) containing multiple cells are analyzed to identify regions with abnormal cell clusters, overlapping cells, and architectural abnormalities. \textbf{WSI-level:} Gigapixel whole-slide images (typically 15,000$\times$15,000 pixels or larger) are analyzed to produce a slide-level diagnosis, either by aggregating patch-level predictions or through end-to-end slide-level models. This hierarchical organization mirrors the clinical workflow, where cytopathologists first scan at low magnification for suspicious regions, then examine individual cells at high magnification for diagnostic classification.

	\subsection{Screening Pathways in the HPV Era}
	\label{sec:hpv_lmic}
Cervical screening is shifting from cytology-alone to HPV-based primary screening. The WHO 90-70-90 strategy recommends HPV testing as the primary screening method, with cytology used for triage of HPV-positive women or as co-testing where resources allow \cite{Who2021human}. This changes the operating point of AI cytology: the population reaching cytology is enriched for HPV-positive and often older women, the prevalence of abnormalities is higher than in unselected screening, and the clinically critical task becomes risk stratification rather than binary normality. Self-sampling further shifts the workflow, because samples may be collected outside a clinic and processed in batches. AI systems should therefore be evaluated on HPV-enriched and self-collected populations, not only on convenience samples from academic centers.

\begin{figure*}[t]
		\centering
		\includegraphics[width=0.95\linewidth]{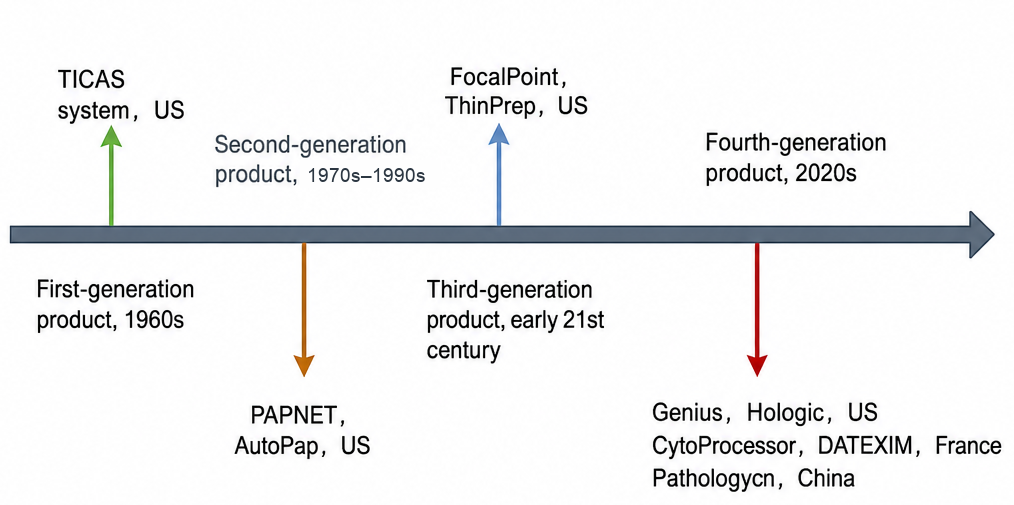}
		\caption{Four generations of commercial cervical cytology screening systems. Generation 1 (1960s): TICAS. Generation 2 (1970s--1990s): PAPNET and AutoPap. Generation 3 (early 2000s): FocalPoint GS and ThinPrep Imaging. Generation 4 (2019--2020s): Hologic Genius, DATEXIM CytoProcessor, and Chinese systems such as Pathologycn.}
		\label{fig:commercial}
	\end{figure*}

The HPV-era pathways described above build on a longer history of automated cervical cytology systems. The core idea is to mimic the cytopathologist's workflow by combining cytopathology knowledge with image processing and, increasingly, deep learning, with the goal of computer-aided diagnosis. Four generations can be distinguished (Fig.~\ref{fig:commercial}). The first dates to the 1960s, when the TICAS system was developed in the United States \cite{Wied1968Taxonomic}. The second comprises PAPNET and AutoPap, developed in the United States in the early 1970s; in 1995, the FDA approved both for assisted diagnostic applications \cite{koss2000application,Patten1996Neopath}. The third includes the FocalPoint GS Imaging and ThinPrep Imaging systems \cite{Kardos2004Focal,2006Evaluation}. Hologic's Genius Digital Diagnostics System received CE marking in 2020 and FDA De Novo authorization in January 2024 \cite{Hologic2022,fda2024genius}; DATEXIM's CytoProcessor received EU certification in 2019 \cite{crowell2019cytoprocessortm}.

In China, automated cervical screening started later but accelerated with the rise of deep learning and increasing demand for cervical screening. Universities and companies---including Pathologycn \cite{kunyuan2021}, Hnpixin \cite{pinxin2021}, KingMed \cite{jinyu2019}, iDeepWise \cite{ideepwise2021}, Tencent Miying \cite{tencent2021}, and Imsight \cite{shijian2021}---have developed deep learning-based cytology systems. Pathologycn's cervical slide scanning and analysis system obtained a Class II medical device registration certificate from Hunan Province and entered the market \cite{kunyuan2021}. Despite this progress, clinical and ethical considerations mean that automated cytology systems remain diagnostic aids rather than autonomous diagnostic devices.

	\subsection{Two Clinical Technical Roadmaps}
	\label{sec:roadmaps}




Two dominant technical roadmaps have emerged for clinical cervical cytology screening:

\textbf{Roadmap 1: WSI Confidence Aggregation.} The WSI is divided into patches or ROIs, each classified by a CNN. Cell-level features or confidence scores are ranked, and the top-K most suspicious cells are selected for slide-level classification via aggregation (averaging, voting, or learned fusion). The screening interface displays the top-K cells sorted by abnormality score, providing a ranked review list but without explicit cell type labels. This approach requires only slide-level or patch-level labels, minimizing annotation cost. However, the lack of cell type information reduces transparency: cytopathologists cannot verify whether the model correctly identifies ASC-US versus HSIL cells, limiting trust in the system's decisions.

\textbf{Roadmap 2: Cell Detection and Classification.} The WSI is partitioned into patches, and an end-to-end detection model (e.g., YOLO3 with cascaded classifiers, as in Zhu et al. \cite{Zhu2021}) simultaneously localizes and classifies each abnormal cell. The screening interface displays detected cells with their Bethesda category labels, enabling cytopathologists to verify individual cell classifications. This approach provides substantially greater transparency and clinical trust. However, it requires dense per-cell bounding box and category annotations, which are expensive to obtain.

The 2024--2026 emergence of label-efficient learning paradigms---semi-supervised detection using unlabeled data \cite{zhang2024scac}, weakly supervised MIL requiring only slide-level labels \cite{fei2025weakly}, and large-scale self-supervised pretraining \cite{jiang2025generalizable}---is progressively reducing the annotation cost of Roadmap 2. Test-time adaptation methods enabling deployment across scanner types and clinical centers without site-specific retraining further accelerate clinical viability. Roadmap 2 represents the developmental trajectory of the field, with the annotation burden being systematically addressed by contemporary learning paradigms.

\section{Review Methodology}
\label{sec:review_methodology}

We used a structured search and screening process; this is a narrative survey rather than a full systematic review or meta-analysis. We searched PubMed, Scopus, Web of Science, IEEE Xplore, and arXiv from January 2010 to October 2026. The search combined cervical cytology terms (``cervical cytology'', ``Pap smear'', ``liquid-based cytology'', ``ThinPrep'', ``TCT''), method terms (``deep learning'', ``convolutional neural network'', ``transformer'', ``foundation model'', ``multiple-instance learning'', ``self-supervised''), and task terms (``segmentation'', ``classification'', ``detection'', ``whole-slide image''). We included peer-reviewed articles and preprints that (i) addressed cervical cytology image analysis, (ii) reported a quantitative evaluation on cytology data, and (iii) described a method, dataset, or benchmark relevant to at least one data scale (cell, patch, or WSI). We excluded records that (i) were not cytology-based (colposcopy, histopathology, or HPV testing without a cytology component), (ii) were of low quality (no quantitative evaluation or insufficient methodological detail), or (iii) were reviews without primary data. English-language publications were included; regulatory documents were retrieved separately from the FDA and NMPA databases.

One author (Y.J.) screened all titles and abstracts and assessed records with sufficient textual information in full text. Data were extracted for study design, data scale, task, dataset, supervision strategy, evaluation metrics, external validation, and code availability. Risk of bias was assessed by the same author with QUADAS-2 for diagnostic accuracy studies and CLAIM for AI studies; reporting completeness was checked against TRIPOD-AI where applicable. Because study designs, datasets, and metrics were heterogeneous, evidence was synthesized qualitatively; no meta-analysis was performed.

The search and screening flow is summarized in Table~\ref{tbl:screening_flow}. Of 163 records identified, one duplicate was removed, leaving 162 records for title/abstract screening. Twenty records were excluded at this stage because they lacked sufficient textual information or were clearly out of scope. The remaining 142 records were assessed in full text; 9 were excluded as non-cytology, 5 as low quality, and 3 because the full text could not be obtained. The final corpus comprised 125 records (primary studies, datasets, and regulatory documents) for qualitative synthesis. Single-author screening and quality assessment are limitations of this review.

	\begin{table}[t] \tiny
		\centering
		\caption{Search and screening flow for this review.}
		\begin{tabular}{|p{8.2cm}|c|}
			\hline
			Stage & Records \\ \hline
			Records identified from databases & 163 \\ \hline
			Duplicates removed & 1 \\ \hline
			Records screened at title/abstract & 162 \\ \hline
			Records excluded at title/abstract & 20 \\ \hline
			Full-text records assessed & 142 \\ \hline
			Excluded: non-cytology & 9 \\ \hline
			Excluded: low quality or insufficient methodological detail & 5 \\ \hline
			Excluded: full text unavailable & 3 \\ \hline
			Records included in qualitative synthesis & 125 \\ \hline
		\end{tabular}
		\label{tbl:screening_flow}
	\end{table}

\section{Taxonomy and Evaluation Framework}
\label{sec:framework}

This section defines the three-axis taxonomy that organizes the literature and the evaluation and reporting framework used to compare methods.

	\subsection{Three-Axis Taxonomy}
	\label{sec:taxonomy}

	\begin{figure}
		\centering
		\includegraphics[width=0.95\linewidth]{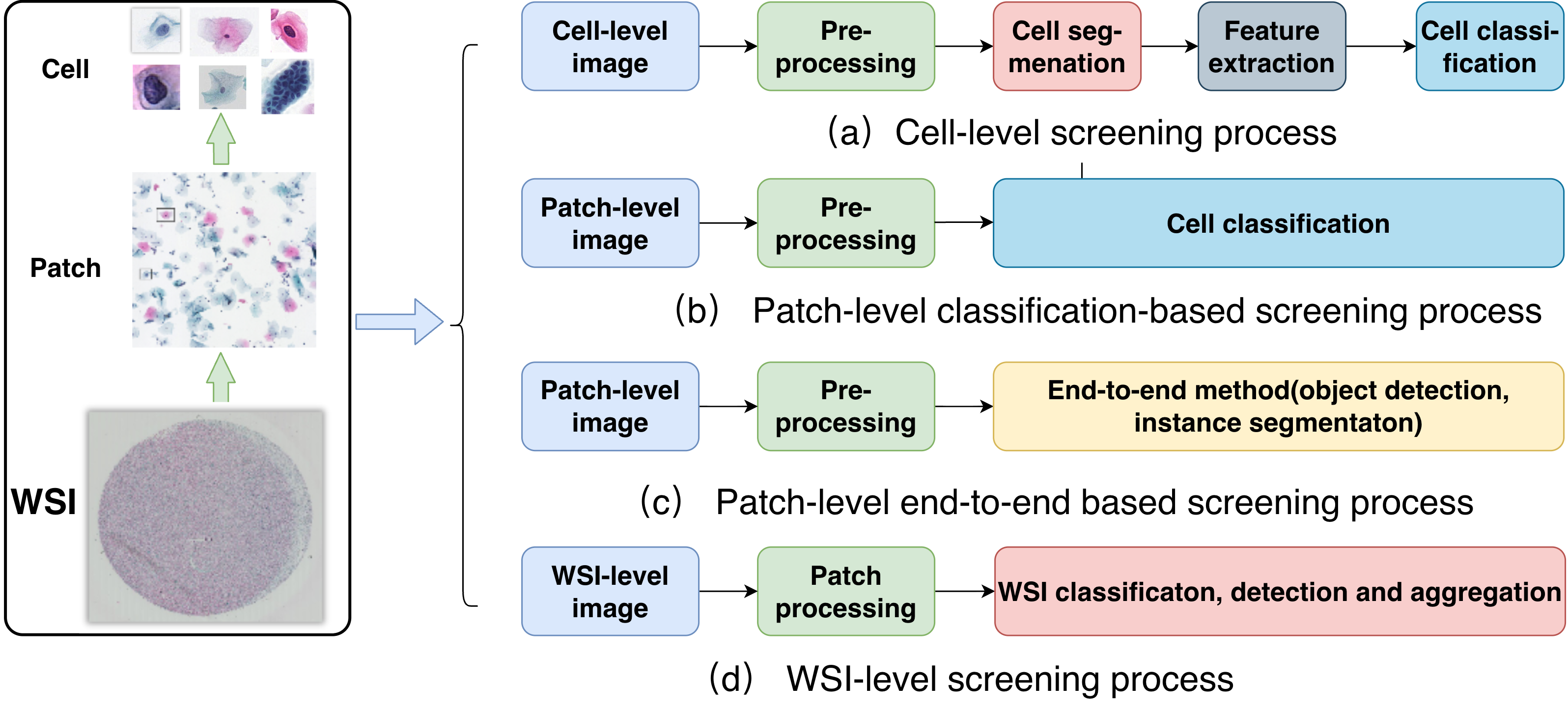}
		\caption{Four processing pipelines for automated cervical screening: cell-level segmentation, feature extraction, and classification; patch-level classification; patch-level end-to-end detection; and WSI-level classification, detection, and aggregation.}
		\label{fig:fig_pipelines}
	\end{figure}



We propose a hierarchical, three-axis taxonomy for cervical cytology screening methods. The primary axis is \textbf{data scale}, reflecting the spatial granularity at which analysis is performed; the secondary axis is \textbf{functional category}, describing the computational task; the third axis is \textbf{supervision}, describing the label regime required for training. This taxonomy captures the technical evolution of the field, the clinical diagnostic hierarchy, and the annotation economics that determine deployability.

\textbf{Data Scale Axis (Primary):}

1. \textbf{Cell-level:} Analysis of individual cervical cells (typically 64$\times$64 to 128$\times$128 pixel crops). Tasks include nucleus and cytoplasm segmentation, morphological feature extraction, and cell-level classification into Bethesda categories. Input scale: single cells. Clinical application: differential diagnosis at cellular resolution.

2. \textbf{Patch-level:} Analysis of image regions containing multiple cells (typically 256$\times$256 to 1024$\times$1024 pixels). Tasks include instance segmentation (delineating and classifying each cell within the patch) and object detection (localizing and classifying abnormal cells via bounding boxes). Input scale: multi-cell image regions. Clinical application: regional abnormality screening where cell context and spatial relationships matter.

3. \textbf{WSI-level:} Analysis of gigapixel whole-slide images. Tasks include slide-level abnormality classification, risk stratification, and lesion localization through aggregation of cell-level or patch-level predictions, or through end-to-end slide-level models. Input scale: 15,000$\times$15,000 pixels or larger. Clinical application: slide-level triage and diagnostic decision support.

\textbf{Functional Category Axis (Secondary):}

\begin{itemize}
	\item \textbf{Segmentation:} Delineation of nucleus and cytoplasm boundaries. Methods span classical approaches (watershed, active contours, graph cuts) to deep learning (U-Net, FCN, Mask R-CNN, Transformer- and Mamba-based architectures).
	\item \textbf{Feature Extraction:} Derivation of discriminative representations. Categories include handcrafted features (morphological: area, perimeter, eccentricity; textural: GLCM, LBP, Haralick; chromatic: brightness, color moments), deep features extracted from pretrained CNNs (VGG, ResNet, Inception), and hybrid features fusing both paradigms.
	\item \textbf{Classification:} Assignment of cells or patches to Bethesda diagnostic categories. Classical classifiers (SVM, ANN, random forest) operate on extracted features, while deep classifiers (CNN, ViT, ensemble methods) learn end-to-end from raw images.
	\item \textbf{Detection:} Simultaneous localization and classification of cells. Approaches include two-stage detectors (Faster R-CNN, Cascade R-CNN), one-stage detectors (SSD, YOLO, RetinaNet), and query-based detectors (DETR variants).
	\item \textbf{Quality Control and ROI Selection:} Detection of unsatisfactory slides, artifacts, staining problems, and selection of diagnostically relevant regions before or alongside diagnosis. This function is often omitted from benchmarks but is critical in deployment.
	\item \textbf{Report Generation and Human-AI Interaction:} Structured diagnostic reports, confidence and uncertainty estimates, and review queues that integrate model outputs into cytopathologist workflow.
\end{itemize}

\textbf{Supervision Axis (Third):} Methods also differ in their label requirements: fully supervised, weakly supervised (slide-level labels), semi-supervised (small labeled plus large unlabeled sets), self-supervised (pretraining without labels), few-shot/zero-shot, and active learning. This axis is clinically decisive because annotation cost, rather than architecture, often determines whether a method can be deployed. We therefore report the supervision regime for each method family in Sections~\ref{sec:cell}--\ref{sec:wsi} and synthesize the paradigms in Section~\ref{sec:paradigms}.

The four processing pipelines are illustrated in Fig.~\ref{fig:fig_pipelines}. The field has evolved through four technical generations. The key technical transition from each generation to the next is driven by a reduction in manual engineering: Generation 1 relies on handcrafted features and classical classifiers; Generation 2 replaces individual pipeline stages with deep networks; Generation 3 eliminates the explicit segmentation step through end-to-end detection; Generation 4 introduces foundation model pretraining and label-efficient learning to address annotation scarcity and domain shift.

\begin{figure}
		\centering
		\includegraphics[width=0.95\linewidth]{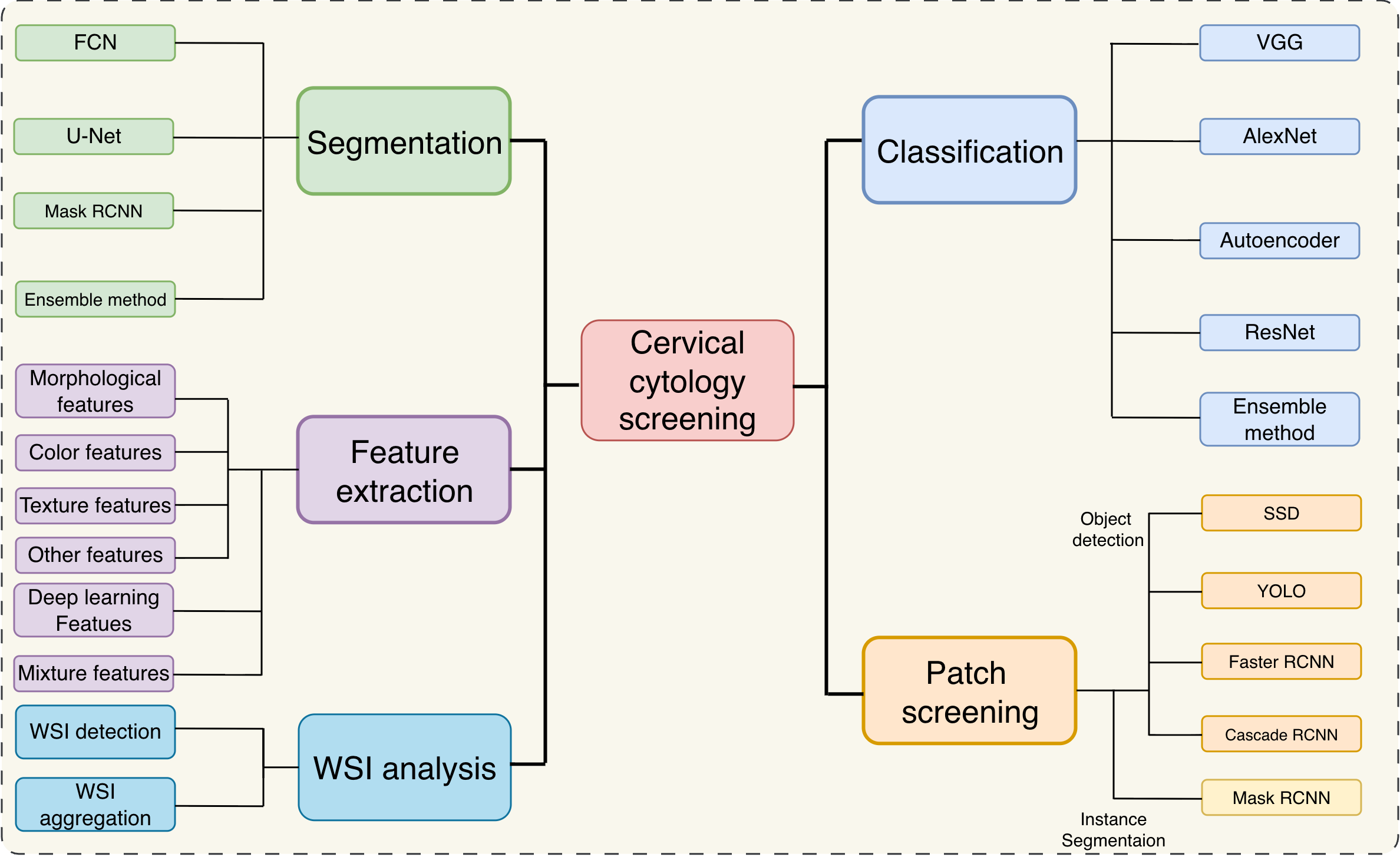}
		\caption{Functional taxonomy of deep learning methods for cervical screening: four core computational branches---segmentation, feature extraction, classification, and detection. Classical feature descriptors are retained under feature extraction; classical segmentation and classification methods are discussed in the text but omitted from the figure. Quality control and report generation are workflow functions discussed in Section~\ref{sec:taxonomy}.}
		\label{fig:fig3}
	\end{figure}

Fig.~\ref{fig:fig3} provides a detailed functional taxonomy of deep learning methods. It focuses on the four core computational branches; classical segmentation and classification methods are discussed in the text but omitted from the figure, and quality control and report generation are workflow functions discussed in Section~\ref{sec:taxonomy}. Segmentation methods have evolved from classical approaches (watershed, level sets, graph cuts) through deep encoder-decoder architectures (U-Net, FCN, DeepLab) to attention-enhanced and state-space models. Feature extraction has progressed from manual engineering of 29-dimensional morphometric vectors to deep features from VGG, ResNet, and Inception families, and more recently to self-supervised representations. Classification has advanced from SVM and ANN operating on handcrafted features to CNN ensembles achieving over 99\% accuracy on curated single-cell benchmarks. Detection has transitioned through two-stage (Faster R-CNN, Cascade R-CNN), one-stage (SSD, YOLO, RetinaNet), and query-based (DETR) paradigms.

\begin{figure}
		\centering
		\includegraphics[width=0.85\linewidth]{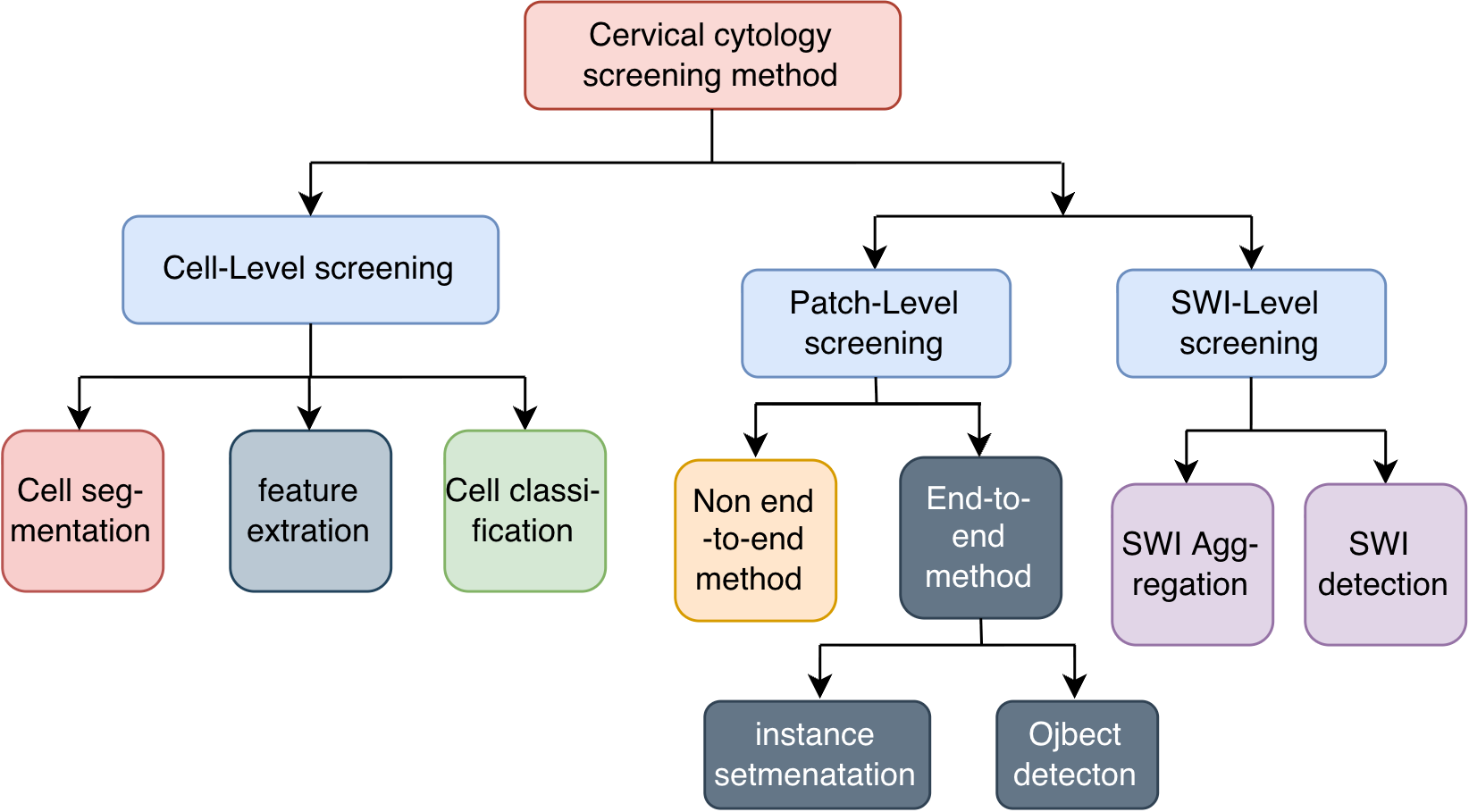}
		\caption{Data-level taxonomy: cell-level segmentation, feature extraction, and classification; patch-level instance segmentation and detection; and WSI-level confidence aggregation and cell detection. Classical branches are shown as background. Supervision is an orthogonal third axis discussed in the text.}
		\label{fig:fig4}
	\end{figure}

Fig.~\ref{fig:fig4} presents the complete data-level taxonomy, the central organizational framework of this survey. The taxonomy is structured in three tiers. \textbf{Tier 1 (Cell-level):} Segmentation methods (classical and deep learning) extract nucleus and cytoplasm boundaries; feature extraction methods derive discriminative representations; classification methods assign Bethesda categories. \textbf{Tier 2 (Patch-level):} Instance segmentation methods simultaneously segment and classify each cell within a patch; object detection methods localize and classify cells via bounding boxes, using two-stage (Faster R-CNN, Cascade R-CNN), one-stage (SSD, YOLO, RetinaNet), or query-based (DETR) architectures. \textbf{Tier 3 (WSI-level):} WSI confidence methods aggregate patch-level or cell-level prediction scores to produce slide-level diagnoses; cell detection methods perform end-to-end detection across the entire slide, providing per-cell localization and classification. This taxonomy reflects both the increasing data scale and the clinical workflow: pathologists first survey the entire slide (WSI-level), then examine suspicious regions (patch-level), and finally classify individual cells at high magnification (cell-level). Fig.~\ref{fig:fig4} visualizes the scale--function plane; supervision is orthogonal to this plane and is discussed in each section. Table~\ref{tbl:tlb11} translates the taxonomy into a deployment-oriented decision matrix.

	\begin{table}[t] \tiny
		\centering
		\caption{Deployment-oriented decision matrix derived from the three-axis taxonomy.}
		\begin{tabular}{|p{2.6cm}|p{1.4cm}|p{1.4cm}|p{3.0cm}|p{3.0cm}|}
			\hline
			Deployment setting & WSI scanner & Label budget & Recommended route & Primary metrics \\ \hline
			High-resource screening laboratory & Yes & High & Roadmap 2 (cell detection) or WSI MIL & Sensitivity at fixed specificity; PPV; turnaround time \\ \hline
			Low-resource, no scanner & No & Low & WSI-free/edge AI (e.g., CYTOLONE) & NPV; offline latency; cost per case \\ \hline
			Rare cell types (AGC, AIS) & Any & Very low & Few-shot/SSL + hierarchical classifier & Per-class recall; AUPRC \\ \hline
			Slide-level labels only & Yes & Low & MIL/weak supervision & AUC; calibration \\ \hline
			Method benchmark & Public data & Medium & Fixed splits + external validation & DSC/AJI/mAP with 95\% CI \\ \hline
		\end{tabular}
		\label{tbl:tlb11}
	\end{table}

	\subsection{Evaluation Metrics}

Automated cervical cytology screening employs task-specific evaluation metrics. \textbf{Classification} is assessed using accuracy, precision, recall, F1 score, sensitivity, and specificity. \textbf{Segmentation} performance is measured through the Dice Similarity Coefficient (DSC), Jaccard index (Intersection over Union), and Aggregated Jaccard Index (AJI) for overlapping cell scenarios. \textbf{Detection} is evaluated using mean Average Precision (mAP) and AP50 (Average Precision at IoU threshold 0.5). For instance segmentation, Panoptic Quality (PQ) provides a unified metric jointly measuring segmentation and recognition quality. \textbf{WSI-level} screening uses the Area Under the ROC Curve (AUC) for slide-level abnormality assessment. These metrics are defined in Table~\ref{tlb:tlb1}, where TP, TN, FP, FN denote true positives, true negatives, false positives, and false negatives respectively, and A and B represent ground truth and predicted regions. Accuracy alone is not sufficient for screening, because abnormality prevalence is low and a model can achieve high accuracy while missing most positive cases; prevalence-aware reporting is therefore required (Section~\ref{sec:reporting}).

	\begin{table}[t] \tiny
		\centering
		\caption{Evaluation metrics for cervical cytology screening tasks.}
		\begin{tabular}{|l|c|p{5cm}|}
			\hline Metric & Formula & Primary Application \\
			\hline Precision & $\frac{TP}{TP+FP}$ & Classification, Detection \\
			\hline Recall & $\frac{TP}{TP+FN}$ & Classification, Detection \\
			\hline F1 score & $2 \times \frac{P \times R}{P+R}$ & Classification (harmonic mean) \\
			\hline Sensitivity  & $\frac{TP}{TP+FN}$ & Classification, Screening \\
			\hline Specificity & $\frac{TN}{TN+FP}$ & Classification, Screening \\
			\hline Accuracy & $\frac{TP+TN}{TP+TN+FP+FN}$ & Classification \\
			\hline Dice (DSC)  & $ \frac{2|A \cap B|}{|A|+|B|} $ & Segmentation \\
			\hline Jaccard (IoU)  & $ \frac{|A \cap B|}{|A| \cup |B|} $ & Segmentation, Detection \\
			\hline mAP & Average precision across categories & Object Detection \\
			\hline AP50  &  AP at IoU = 0.5 & Object Detection \\
			\hline AUC & Area under ROC curve & WSI-level Screening \\
			\hline PQ & $\underbrace{\frac{|TP|}{|TP|+\frac{1}{2}|FP|+\frac{1}{2}|FN|}}_{\text{Recognition quality}} \times \underbrace{\frac{\sum\text{IoU}}{|TP|}}_{\text{Segmentation quality}}$ & Instance Segmentation \\
			\hline
		\end{tabular}
		\label{tlb:tlb1}
	\end{table}

	\subsection{Reporting Standards and Statistical Comparison}
	\label{sec:reporting}
Metrics are necessary but not sufficient: in screening settings, the prevalence of abnormal cells or slides is low (typically 3--8\%), so a high accuracy can coexist with poor positive predictive value. We therefore recommend that studies report a minimum set of prevalence-aware measures: sensitivity at fixed specificity (e.g., 90\% and 95\%), PPV and NPV at a stated prevalence, the full ROC curve and AUC with 95\% confidence intervals, and calibration (reliability diagram and expected calibration error). For segmentation and detection, authors should state the metric definition and IoU threshold, and report per-class results for clinically critical categories (ASC-US, ASC-H, AGC, AIS) rather than only aggregate scores.

Comparisons should use patient-level or slide-level splits, and must state whether patches from the same slide or patient appear in both training and test sets. Paired comparisons should use McNemar's test for classification and paired bootstrap for detection or segmentation metrics; AUC differences should be tested with DeLong's method. External validation should report the number of centers, scanners, staining protocols, and the target population. We recommend CLAIM, TRIPOD-AI, and STARD-AI for reporting, and QUADAS-2 for risk-of-bias assessment. Table~\ref{tlb:tlb2} summarizes the minimum reporting checklist.

	\begin{table}[t] \tiny
		\centering
		\caption{Minimum reporting checklist for cervical cytology AI studies.}
		\begin{tabular}{|p{3.2cm}|p{8.2cm}|}
			\hline Item & Requirement \\ \hline
			Data split & Patient-level or slide-level; no patch leakage across training and test sets \\ \hline
			Prevalence & Report the abnormal rate in the evaluation cohort \\ \hline
			Classification & Sensitivity, specificity, AUC (95\% CI); sensitivity at fixed specificity; PPV/NPV at stated prevalence \\ \hline
			Calibration & Reliability diagram and expected calibration error (ECE) \\ \hline
			Segmentation & DSC/AJI with explicit definition; per-class or per-cell-type results \\ \hline
			Detection & mAP with explicit IoU threshold; AP50 and AP75; FROC when clinically relevant \\ \hline
			Uncertainty & 95\% confidence intervals for all primary metrics \\ \hline
			Statistical test & McNemar (classification), paired bootstrap (detection/segmentation), DeLong (AUC) \\ \hline
			External validation & Number of centers, scanners, stains, and population; prospective vs retrospective \\ \hline
			Reporting & CLAIM, TRIPOD-AI, STARD-AI; QUADAS-2 risk of bias \\ \hline
		\end{tabular}
		\label{tlb:tlb2}
	\end{table}

Tables~\ref{tlb:abbrevA} and~\ref{tlb:abbrevB} summarize the clinical, technical, and reporting abbreviations used in this survey.

	\begin{table}[t] \scriptsize
		\centering
		\caption{Clinical, pathological, and imaging abbreviations used in this survey.}
		\begin{tabular}{|p{2.3cm}|p{3.65cm}|p{2.3cm}|p{3.65cm}|}
			\hline
			Abbrev. & Meaning & Abbrev. & Meaning \\ \hline
			TBS & Bethesda System & WSI & Whole-slide image \\ \hline
			ASC-US & Atypical squamous cells of undetermined significance & ROI & Region of interest \\ \hline
			ASC-H & Atypical squamous cells, cannot exclude HSIL & CAD & Computer-aided diagnosis/detection \\ \hline
			LSIL & Low-grade squamous intraepithelial lesion & HPV & Human papillomavirus \\ \hline
			HSIL & High-grade squamous intraepithelial lesion & TCT & ThinPrep cytologic test \\ \hline
			SCC & Squamous cell carcinoma & LBC & Liquid-based cytology \\ \hline
			AC & Adenocarcinoma & EDF & Extended depth of field \\ \hline
			AGC & Atypical glandular cells & N/C & Nuclear-to-cytoplasmic ratio \\ \hline
			AIS & Adenocarcinoma in situ & LMIC & Low- and middle-income country \\ \hline
			NILM & Negative for intraepithelial lesion or malignancy & WHO & World Health Organization \\ \hline
		\end{tabular}
		\label{tlb:abbrevA}
	\end{table}

	\begin{table}[t] \scriptsize
		\centering
		\caption{AI, evaluation, regulatory, and reporting abbreviations used in this survey.}
		\begin{tabular}{|p{2.3cm}|p{3.65cm}|p{2.3cm}|p{3.65cm}|}
			\hline
			Abbrev. & Meaning & Abbrev. & Meaning \\ \hline
			AI & Artificial intelligence & MIL & Multiple-instance learning \\ \hline
			CNN & Convolutional neural network & SSL & Self-supervised learning \\ \hline
			FCN & Fully convolutional network & TTA & Test-time adaptation \\ \hline
			R-CNN & Region-based convolutional neural network & PEFT & Parameter-efficient fine-tuning \\ \hline
			YOLO & You Only Look Once & FM & Foundation model \\ \hline
			DETR & Detection Transformer & AUC & Area under the ROC curve \\ \hline
			RNN & Recurrent neural network & ROC & Receiver operating characteristic \\ \hline
			ViT & Vision Transformer & TP/TN/FP/FN & True/false positives/negatives \\ \hline
			CLIP & Contrastive Language--Image Pretraining & DSC & Dice similarity coefficient \\ \hline
			LLM/LVLM & Large language model / large vision-language model & AJI & Aggregated Jaccard index \\ \hline
			QA/VQA & Question answering / visual question answering & ZSI & Zijdenbos similarity index \\ \hline
			PQ & Panoptic quality & mAP & Mean average precision \\ \hline
			AP50/AP75 & Average precision at IoU 0.50/0.75 & FROC & Free-response ROC \\ \hline
			PPV/NPV & Positive/negative predictive value & ECE & Expected calibration error \\ \hline
			CI & Confidence interval & F1 & Harmonic mean of precision and recall \\ \hline
			IoU & Intersection over union & CLAIM & Checklist for AI in Medical Imaging \\ \hline
			QUADAS-2 & Quality Assessment of Diagnostic Accuracy Studies 2 & TRIPOD-AI & Transparent Reporting of a multivariable prediction model for Individual Prognosis Or Diagnosis---AI \\ \hline
			STARD-AI & Standards for Reporting of Diagnostic Accuracy Studies---AI & FDA & U.S. Food and Drug Administration \\ \hline
			NMPA & National Medical Products Administration & CE & Conformit\'e Europ\'eenne \\ \hline
			TIS & ThinPrep Imaging System & QALY & Quality-adjusted life year \\ \hline
			N/R & Not reported & N/A & Not applicable \\ \hline
			pp & Percentage points & & \\ \hline
		\end{tabular}
		\label{tlb:abbrevB}
	\end{table}

\section{Datasets and Benchmarks}
\label{sec:datasets}











Public cervical cytology datasets remain limited in scale, diversity, and annotation granularity relative to natural image benchmarks. The most widely used benchmarks---Herlev (917 cells, 2005) and SIPaKMeD (4,049 cells, 2018)---are orders of magnitude smaller than ImageNet (1.4M images). Table~\ref{tlb:tlb0} compares 16 publicly available or publicly documented cytology datasets spanning cell-level to WSI-level scales. CAISHI, a histopathology dataset for adenocarcinoma in situ, is discussed separately because it is not a cytology benchmark.

The \textbf{Herlev dataset} \cite{jantzen2005pap} (917 single-cell images, 2005) provides nucleus and cytoplasm segmentation masks and 7-class Bethesda labels. It remains the most widely used benchmark for cell-level segmentation and classification, despite its small size and single-center origin. The \textbf{ISBI 2014/2015 challenge datasets} \cite{Lu2015,Lu2017} contain overlapping cervical cells in Extended Depth of Field (EDF) format---ISBI14 includes 16 EDFs with 945 synthetic patches, ISBI15 includes 17 EDFs with 340 fields of view---with pixel-level nucleus and cytoplasm annotations. \textbf{CERVIX93} \cite{phoulady2018new} adds 93 EDFs (approximately 1,860 fields of view at 20 fields per EDF) with nucleus-only segmentation labels.

The \textbf{CRIC dataset} \cite{cric_cervix2020} provides 400 patches and 3,233 annotated cells with 6-class Bethesda labels, bridging cell-level and patch-level analysis. The \textbf{SIPaKMeD dataset} \cite{Plissiti2018} contains 996 patches and 4,049 single-cell images with 5-class labels, designed for patch-level classification. The \textbf{Tianchi dataset} \cite{tianchi2021} (1,000 WSIs from 2019, with 800 training and 200 test slides) is among the first public WSI-level benchmarks, providing ROI annotations for abnormal regions. The \textbf{ComparisonDetector dataset} \cite{liang2018comparison} includes 6,666 patch-level images with 11-class labels for cell detection.

Several recent datasets address specific limitations. \textbf{Cx22} (1,320 images) provides full instance-level annotations---bounding boxes, segmentation masks, and category labels---for every cell, enabling both detection and instance segmentation \cite{liu2022cx22}. \textbf{HiCervix} (40,229 cells, 29-class hierarchy) is the largest annotated single-cell cervical dataset, with a three-level hierarchical classification scheme and both nuclear and cytoplasmic boundaries \cite{cai2024hicervix}. \textbf{APACC} provides 103,675 annotated cell images from 107 conventional Pap smear whole-slide images, with labels from three expert cytopathologists \cite{kupas2024annotated}. \textbf{RIVA} contains 959 conventional Pap smear patches (15,949 unique cells) with up to four independent expert annotations per cell, enabling inter-observer agreement analysis \cite{perezbianchi2025riva}. \textbf{Zhang et al.} released 8,037 annotated TCT images from 129 slides specifically for detection benchmarking \cite{zhang2025large}. \textbf{SCAC} provides a large unlabeled cervical cytology collection for semi-supervised abnormal-cell detection \cite{zhang2024scac}. \textbf{CCS-127K} (127,471 WSIs from 48 clinical centers) is the largest WSI-level cervical screening collection \cite{jiang2025generalizable}. \textbf{CCBench} (773 QA + 420 VQA pairs) provides a benchmark for evaluating vision-language models on cervical cytology tasks \cite{hong2025evaluating}. \textbf{CAISHI} provides 2,240 H\&E-stained histopathology images for adenocarcinoma in situ; because it is not a cytology dataset, it is discussed separately rather than pooled with cytology benchmarks \cite{yang2024caishi}.

Despite this progress, a critical gap persists: no large-scale, multi-center, publicly accessible WSI-level benchmark with standardized train/validation/test splits and longitudinal clinical outcomes exists for cervical cytology---analogous to TCGA in cancer genomics or CAMELYON in digital pathology. APACC, RIVA, and the TCT detection set partially address the scale and annotation gaps, but they remain single-country collections without outcome labels or standardized external test splits \cite{kupas2024annotated,perezbianchi2025riva,zhang2025large}. The construction of a multi-center benchmark with fixed splits and longitudinal outcomes remains an urgent priority for reproducible and clinically meaningful research.

	\begin{table*} \tiny
		\caption{Summary of 16 publicly available or publicly documented cervical cytology datasets. ``Scale'' gives the analysis level and dataset size; ``Cytology type'' distinguishes conventional Pap smears, liquid-based/TCT, and extended-depth-of-field (EDF) data. ``Annotation'' indicates the label and mask types; ``Annotators/agreement'' reports the number of independent expert annotators when available. N/R = not reported in the source publication; the frequent N/R entries highlight the need for standardized dataset datasheets. CAISHI, a histopathology dataset for adenocarcinoma in situ, is discussed separately and is not included in this table.} %
		\centering
		\resizebox{\textwidth}{!}{%
		\begin{tabular}{|p{2.2cm}|c|p{2.6cm}|p{1.9cm}|p{2.4cm}|p{2.3cm}|p{2.2cm}|p{2.5cm}|}
			\hline
			Dataset & Year & Scale (type; size) & Cytology type & Annotation & Annotators & Primary task & Key limitation \\ \hline
			Herlev \cite{jantzen2005pap} & 2005 & Cell; 917 cells & Conventional Pap & 7-class; N/C masks & N/R & Seg., Cls. & Small; single center \\ \hline
			ISBI14 \cite{Lu2015} & 2014 & Patch; 16 EDFs, 945 patches & EDF & N/C masks & N/R & Seg. & Synthetic patches \\ \hline
			ISBI15 \cite{Lu2017} & 2015 & Patch; 17 EDFs, 340 FOVs & EDF & N/C masks & N/R & Seg. & Single staining \\ \hline
			CERVIX93 \cite{phoulady2018new} & 2017 & Patch; 93 EDFs, 1,860 FOVs & EDF & Nucleus masks & N/R & Seg. & No cytoplasm \\ \hline
			CRIC \cite{cric_cervix2020} & 2020 & Patch; 400 patches, 3,233 cells & N/R & 6-class & N/R & Cls. & No masks \\ \hline
			SIPaKMeD \cite{Plissiti2018} & 2018 & Cell/patch; 996 patches, 4,049 cells & Conventional Pap & 5-class & N/R & Cls. & No localization \\ \hline
			Tianchi \cite{tianchi2021} & 2019 & WSI; 1,000 WSIs & N/R & ROI, 2-class & N/R & Det. & Coarse ROI \\ \hline
			ComparisonDetector \cite{liang2018comparison} & 2018 & Patch; 6,666 patches & N/R & 11-class boxes & N/R & Det. & Single center \\ \hline
			Cx22 \cite{liu2022cx22} & 2022 & Patch; 1,320 images & N/R & Box+mask+class & N/R & Inst. seg., Det. & Moderate scale \\ \hline
			HiCervix \cite{cai2024hicervix} & 2024 & Cell; 40,229 cells, 4,496 WSIs & TCT & 29-class hierarchy; N/C masks & N/R & Hier. seg., Cls. & Single center \\ \hline
			APACC \cite{kupas2024annotated} & 2024 & Patch; 103,675 cells, 107 WSIs & Conventional Pap & Cell labels & 3 cytopathologists & Cls., Det. & Single country \\ \hline
			RIVA \cite{perezbianchi2025riva} & 2025 & Patch; 959 patches, 15,949 cells & Conventional Pap & 8-class; center points & Up to 4 pathologists & Cls., agreement & No outcomes \\ \hline
			TCT detection \cite{zhang2025large} & 2025 & Patch; 8,037 images, 129 slides & TCT & Detection boxes & N/R & Det. & Single center \\ \hline
			SCAC \cite{zhang2024scac} & 2024 & Patch; large unlabeled & N/R & None & N/A & Semi-sup. Det. & No labels \\ \hline
			CCS-127K \cite{jiang2025generalizable} & 2025 & WSI; 127,471 WSIs & N/R & Slide-level & N/R & WSI screening & Restricted access \\ \hline
			CCBench \cite{hong2025evaluating} & 2025 & WSI/QA; 773 QA + 420 VQA & N/R & QA/VQA pairs & N/R & LVLM eval. & Task-specific \\ \hline
		\end{tabular}}
		\label{tlb:tlb0}
	\end{table*}

	\section{Cell-level Methods}
	\label{sec:cell}

We analyze cell-level cervical screening methods according to three functional categories: segmentation methods (delineating nucleus and cytoplasm), feature extraction methods (deriving discriminative representations), and classification methods (assigning Bethesda diagnostic categories).

	\subsection{Segmentation Methods}


Cervical cell segmentation delineates the nucleus and cytoplasm boundaries from microscopy images, forming the foundation for downstream quantitative analysis and classification. Accurate segmentation is challenging due to high cell density, severe overlap and occlusion, substantial variation in cell morphology across the Bethesda spectrum, and inconsistent staining across samples.

We categorize segmentation methods into three classes: (1) classical unsupervised methods (watershed, active contours, graph cuts, thresholding, superpixel-based), (2) deep learning methods (U-Net, FCN, Mask R-CNN, and more recent Transformer and Mamba architectures), and (3) hybrid methods combining deep feature extraction with classical refinement. Classical methods are included for context; the remainder of this section focuses on deep learning architectures. Table~\ref{tbl:tlb3} presents a quantitative comparison.



Deep convolutional neural networks have substantially advanced cervical cell segmentation. Early approaches adapted generic architectures; recent work has introduced cervical-specific designs.


\textbf{U-Net Variants:} Zhao et al. \cite{zhao2019pgu} proposed PGU-net+, a progressive U-Net that learns shape residuals between coarse and fine annotation scales, achieving a DSC of 0.925 and precision of 0.901 on Herlev. Mahyari et al. \cite{Mahyari2021} designed URestNet, a two-stage U-Net with residual blocks and skip connections for iterative refinement. Wan et al. \cite{Wan2019} combined TernausNet for single-cell extraction with DeepLabV2 for cytoplasm segmentation. Ke et al. \cite{Ke2019} used a U-Net with VGG16 encoder and HSDC (Hybrid Standard and Dilated Convolution) decoder modules for joint nucleus-cytoplasm-background segmentation. Zhao et al. \cite{zhao2019automated} introduced D-MEM (Deformable Multi-path Ensemble Model), employing U-Net with dense blocks as a backbone, achieving recall of 0.968.

\textbf{FCN:} Harangi et al. \cite{Harangi2019} ensembled FCN-8 and FCN-16 architectures for cervical cytology segmentation, achieving a DSC of 0.93 on single cells but only 0.534 IoU on multi-cell images, highlighting the difficulty of overlapping cell segmentation. Sabeena et al. \cite{Sabeena2020} applied FCN-based approaches achieving 86.67\% accuracy on multi-cell segmentation.



\textbf{Mask R-CNN:} Kurnianingsih et al. \cite{Kurnianingsih2019} applied Mask R-CNN for contour-level segmentation without classification, achieving precision of 0.92 and recall of 0.91. Zhou et al. \cite{zhou2019irnet} proposed IR-NET, a Mask R-CNN variant incorporating an instance relation module with self-attention for inter-cell information transfer and a sparse-constrained duplicate removal module. IR-NET achieved cytoplasmic AJI of 0.7185 and nuclear AJI of 0.5496 on multi-cell images, underscoring the challenges of overlapping cell segmentation.


\textbf{Other Architectures:} Yang et al. \cite{Yang2020} introduced the Interactive Convolutional Pyramid Network (ICPN) with dual aggregation and selection paths, using Interacting Convolutional Modules (ICM) and Internal Pyramid Resolution Complementing Modules (IPRCM) to capture nuclear size and shape variations.

\textbf{Recent Advances:} The segmentation landscape has expanded beyond CNN encoder--decoder architectures. Recent reviews document growing use of attention-enhanced U-Nets, transformer variants, and hybrid designs \cite{fang2024systematic,valles2026application}. State-space models have also been applied to cervical nuclei and cytoplasm segmentation; a recent VMamba-based model targets accurate boundary delineation, motivated by the linear complexity of SSMs relative to attention \cite{zhang2026vmamba}. However, most recent reports evaluate on different datasets, class definitions, and splits, so their reported Dice values are not directly comparable. Standardized overlapping-cell benchmarks remain the main bottleneck for measuring progress.


\textbf{Hybrid Methods:} Several approaches combine deep feature extraction with classical refinement. Song et al. \cite{song2014deep,song2015accurate,Song2016} integrated multi-scale CNNs with superpixel segmentation (SLIC) and Graph-Cut for cytoplasm and nucleus segmentation, and with level set methods for boundary refinement. Wang et al. \cite{Wang2018} combined SLIC with CNN for initial nucleus segmentation followed by GVF-Snake boundary refinement. Zhang et al. \cite{Zhang2017} combined FCN with graph model methods for nucleus segmentation. Liu et al. \cite{Liu2018} used Mask R-CNN for coarse segmentation followed by locally fully connected conditional random fields for boundary refinement, achieving ZSI of 0.95 on Herlev. Tareef et al. \cite{Tareef2017} proposed superpixel CNNs with cell shape priors for overlapping cell segmentation, achieving nuclear ZSI of 0.94. Zhang et al. \cite{Zhang2020} introduced Attention U-Net with graph-based Random Walk post-processing, achieving DSC of 0.917 on multi-cell images.

	\begin{center}\tiny

		\begin{longtable}{|p{2.5cm}|p{1.5cm}|p{1.8cm}|p{1.8cm}|p{3cm}|}
			\caption{Quantitative comparison of deep learning-based cervical cell segmentation methods. DSC = Dice Similarity Coefficient; AJI = Aggregated Jaccard Index; ZSI = Zijdenbos Similarity Index.} \\

			\hline Method & Year & Architecture & Dataset & Key Performance \\
			\hline U-Net (PGU-net+) \cite{zhao2019pgu} & 2019 & Progressive U-Net & Herlev & DSC 0.925 (ZSI), Prec 0.901 \\
			\hline URestNet \cite{Mahyari2021} & 2021 & Two-stage U-Net+Residual & Private & Single cell segmentation \\
			\hline TernausNet+DeepLab \cite{Wan2019} & 2019 & U-Net+DeepLabV2 & Private & Cell+cytoplasm \\
			\hline VGG16+HSDC U-Net \cite{Ke2019} & 2019 & U-Net+HSDC decoder & Private & 3-class (nuclei, cyto, bg) \\
			\hline D-MEM \cite{zhao2019automated} & 2019 & Deformable U-Net ensemble & Herlev, private & DSC 0.925, Recall 0.968 \\
			\hline FCN ensemble \cite{Harangi2019} & 2019 & FCN-8+FCN-16 ensemble & Herlev & DSC 0.93 (single), IoU 0.534 (multi) \\
			\hline Mask R-CNN \cite{Kurnianingsih2019} & 2019 & Mask R-CNN & Private & Prec 0.92, Recall 0.91 \\
			\hline IR-NET \cite{zhou2019irnet} & 2019 & Mask R-CNN+self-attention & ISBI & AJI 0.72 (cyto), 0.55 (nuclei) \\
			\hline ICPN \cite{Yang2020} & 2020 & Interactive pyramid CNN & Private & Multi-scale nuclear seg. \\
			\hline CNN+Superpixel \cite{song2014deep,Song2016} & 2014--16 & Multi-scale CNN+GraphCut & ISBI14 & Hybrid seg. pipeline \\
			\hline Shape prior+CNN \cite{Tareef2017} & 2017 & Superpixel CNN+shape prior & ISBI14 & ZSI 0.90 (cyto), 0.94 (nuclei) \\
			\hline Mask R-CNN+CRF \cite{Liu2018} & 2018 & Mask R-CNN+LFCCRF & ISBI14 & ZSI 0.95, Prec 0.96 \\
			\hline AttU-Net+RW \cite{Zhang2020} & 2020 & Attention U-Net+Random Walk & ISBI14,15 & DSC 0.917 (multi-cell) \\
			\hline
		\end{longtable}
		\label{tbl:tlb3}
	\end{center}



Classical unsupervised segmentation approaches (watershed, active contours, thresholding) perform adequately on isolated single cells but degrade substantially on overlapping, densely clustered cells. Deep learning methods achieve DSC values exceeding 0.93 on single-cell benchmarks but remain supervised, requiring substantial pixel-level annotation. Single-cell methods commonly report DSC above 0.93, whereas overlapping-cell methods report AJI in the 0.55--0.72 range; because DSC and AJI are different metrics, these values should not be interpreted as a direct effect size. Nevertheless, the consistently lower AJI on overlapping cells shows that multi-cell segmentation remains an open challenge. The recent trend toward Transformer- and Mamba-based architectures with explicit occlusion and overlap modeling, combined with hybrid approaches that pair deep feature extraction with classical shape priors, represents the most promising direction for closing this performance gap.

	\subsection{Feature Extraction}


Effective classification of cervical cells depends on extracting discriminative features that capture clinically relevant morphological alterations. Feature extraction approaches fall into three categories: handcrafted features, deep learning-based features, and hybrid fusion. We summarize handcrafted features briefly as background; the main focus is deep representation learning.

\textbf{Handcrafted Features:} Classical morphometric features quantify the structural changes characteristic of cervical neoplasia. \textit{Morphological features} include area, perimeter, roundness, eccentricity, elongation, compactness, nuclear-to-cytoplasmic (N/C) ratio, nuclear relative position, and long/short diameters of both nucleus and cytoplasm---the N/C ratio being the single most diagnostic morphological parameter in the Bethesda System. \textit{Color features} encompass brightness, gray value, color intensity, minimum/maximum brightness, average color value, N/C brightness ratio, saturation, and optimal density. \textit{Texture features} include LBP, GLCM, Haralick features, Laws texture energy, Tamura features, HoG, MRF, roughness, histogram statistics, and frequency-domain texture and energy descriptors. Table~\ref{tlb:tlb4} provides a comprehensive taxonomy of handcrafted features organized by category and cellular compartment.



\textbf{Deep Learning Features:} CNNs pretrained on ImageNet serve as effective feature extractors for cervical cells through transfer learning. Gv et al. \cite{Gv2019} extracted features from ResNet34 to train an SVM classifier, achieving 99.63\% accuracy on 5-class SIPaKMeD classification. Mousser et al. \cite{Mousser2019} compared features from VGG16, VGG19, ResNet50, and InceptionV3, finding that deeper architectures capture increasingly abstract representations of nuclear morphology. Khamparia et al. \cite{Khamparia2020} evaluated InceptionV3, VGG19, SqueezeNet, and ResNet50 features. Plissiti et al. \cite{Plissiti2018} used VGG19 features for SIPaKMeD classification. Rahaman et al. \cite{Rahaman2021} proposed a deep feature fusion network concatenating features from VGG16, VGG19, ResNet50, and XceptionNet, achieving 99.14\% accuracy on 5-class classification. Nguyen et al. \cite{Nguyen2019,Nguyen2018} fused InceptionV3, ResNet152, and InceptionResNetV2 features.


\textbf{Hybrid Feature Fusion:} Combining handcrafted and deep features leverages complementary strengths---the clinical interpretability of morphometric features and the representational power of deep features. Dong et al. \cite{DongyaoJia2020} combined GLCM, Gabor, and CNN features. Dong et al. \cite{Dong2020} fused InceptionV3 features with color, texture, and morphological features. Rohmatillah et al. \cite{Rohmatillah2018} combined LeNet features with LDA and PCA projections.



Deep learning has eliminated manual feature engineering from state-of-the-art pipelines. However, handcrafted features retain value for two reasons: they are inherently interpretable, mapping directly to Bethesda System diagnostic criteria (N/C ratio, nuclear membrane irregularities, chromatin texture), and they complement deep features in hybrid approaches, particularly for rare cell types where deep models overfit. The principal limitation of deep feature extraction is domain shift: features from ImageNet-pretrained networks may not optimally represent the distinctive texture and chromatic characteristics of Papanicolaou-stained cytology images. Self-supervised pretraining on cervical-specific image collections represents a promising avenue for learning domain-adapted feature representations.


	\begin{table}[t]\tiny
		\centering
		\caption{Comprehensive taxonomy of cervical cell features. N and C denote features computed for nucleus and cytoplasm respectively. Features are organized by category: morphological (shape), color (chromatic), texture (statistical and structural), deep learning-based (CNN-extracted), and other (transform-domain and hybrid).}
		\begin{tabular}{|p{2.0cm}|p{1.2cm}|p{1.5cm}|p{2cm}| p{2cm}| }
			\hline Shape features & Color features & Texture features  & Deep learning-based features  & Other features \\
			\hline area (N,C) & brightness (N,C) & LBP  &  Lenet & fourier operator\\
			\hline roundness (N,C) & gray value (N,C) & GLCM  & Inception  & temporal \\
			\hline eccentricity (N,C) & color intensity & roughness   & ResNet & weighted PCA \\
			\hline long diameter (N,C) & minimum brightness & histogram  & VGG & wavelet \\
			\hline short diameter (N,C) & maximum brightness &  TAS  & Xception-Net  & texture and energy in frequency domain\\
			\hline perimeter (N,C) & average color value & Laws    & Squeeze-Net  & \\
			\hline elongation (N,C) & N/C brightness ratio & moment   &GLCM, Gabor, and CNN   & \\
			\hline compactness & optimal density & Haralick  &  Inception, color, texture, and morphology  &  \\
			\hline maximum value (N,C) & saturation  & mean of red, blue, and green   & Lenet and LDA, PCA   & \\
			\hline minimum value (N,C) &  & Tamura   &  &  \\
			\hline nuclear-cytoplasmic ratio &  & HoG   &  &    \\
			\hline nuclear relative position &  & MRF   &  &    \\
			\hline
		\end{tabular}
		\label{tlb:tlb4}
	\end{table}

	\subsection{Classification}


Cell-level classification assigns individual cervical cells to Bethesda System diagnostic categories. This task represents the final and most clinically consequential step in the cell-level analysis pipeline. Table~\ref{tbl:tlb6} provides a comprehensive quantitative comparison of deep learning classifiers.



Deep learning classifiers for cervical cells predominantly employ CNNs. The key advantage over classical approaches is the elimination of error-prone manual feature engineering: CNNs learn discriminative morphological, textural, and chromatic features directly from raw pixel data. Classical classifiers on handcrafted features are covered by earlier surveys and are not reviewed in detail here.



\textbf{VGG-based Classifiers:} VGG16 and VGG19 have been extensively evaluated. Tripathi et al. \cite{Tripathi2021} achieved 94.38\% precision on 5-class Herlev classification. Kurnianingsih et al. \cite{Kurnianingsih2019} reported 95\% accuracy on 7-class classification. Chen et al. \cite{chen2021cytobrain} proposed CytoBrain, a VGG-based system for compact cell representation. Wieslander et al. \cite{Wieslander2017} applied VGG16 with data augmentation.

\textbf{AlexNet-based Classifiers:} AlexNet provides a computationally efficient baseline. The DeepPap system \cite{Zhang2017} achieved strong results using AlexNet with transfer learning. Yilmaz et al. \cite{Yilmaz2020} reported 93\% accuracy, 96\% precision, and 95\% F1-score on 7-class Herlev classification. Shanthi et al. \cite{Shanthi2019}, Plissiti et al. \cite{Plissiti2018}, and Haryanto et al. \cite{Haryanto2020} also benchmarked AlexNet variants.

\textbf{ResNet-based Classifiers:} ResNet architectures achieve the strongest single-model performance. Gv et al. \cite{Gv2019} reported 99.63\% accuracy with 99.83\% specificity on 5-class SIPaKMeD. Shanthi et al. \cite{Shanthi2019}, Tripathi et al. \cite{Tripathi2021}, and Pirovano et al. \cite{Pirovano2019} further validated ResNet variants across datasets, with consistent accuracy above 95\%.

\textbf{Autoencoder-based Methods:} Adem et al. \cite{Adem2019} applied stacked autoencoders achieving 97.8\% accuracy on Herlev. Wang et al. \cite{Wang2017} explored autoencoder-based feature representations.



\textbf{Ensemble Methods:} Ensemble approaches consistently outperform single models. Rahaman et al. \cite{Rahaman2021} fused features from VGG16, VGG19, ResNet50, and XceptionNet, achieving 98.32\% accuracy on 2-class and 99.14\% on 5-class classification. Nguyen et al. \cite{Nguyen2018,Nguyen2019} ensembled InceptionV3, ResNet152, and InceptionResNetV2. Manna et al. \cite{Manna2021} integrated InceptionV3, Xception, and DenseNet-169. Diniz et al. \cite{Diniz2021} combined EfficientNet, MobileNet, XceptionNet, and InceptionNetV3. Hussain et al. \cite{Hussain2020} ensembled ResNet-50, ResNet-101, and GoogLeNet. Mousser et al. \cite{Mousser2019} integrated VGG16, VGG19, ResNet50, and InceptionV3.

\textbf{Multi-cell Classification:} Several studies classify multi-cell image patches directly. Wang et al. \cite{Wang2020,Wang2021} applied attention mechanisms to multi-cell CNN classification. Gv et al. \cite{Gv2019} and Manna et al. \cite{Manna2021} demonstrated that CNNs can classify multi-cell images, though performance degrades relative to single-cell classification due to cellular heterogeneity and overlapping morphologies.

\textbf{Recent Advances in Classification:} Recent work has explored attention-enhanced CNN variants, ensemble stacking, vision transformers, and contrastive self-supervised pretraining \cite{valles2026application}. Reported accuracies on Herlev and SIPaKMeD are now consistently in the high 90s, but these benchmarks contain fewer than 1,000 cells and are frequently split randomly, so small test sets and near-duplicate cells can dominate differences of a few percentage points. The more informative recent development is the release of hierarchical, multi-class benchmarks such as HiCervix, which evaluates coarse- and fine-grained Bethesda categories and reports substantially lower performance than saturated binary benchmarks \cite{cai2024hicervix}. We therefore treat single-cell accuracy on Herlev/SIPaKMeD as a sanity check rather than evidence of clinical readiness.

	\begin{center} \tiny
		\begin{longtable}{|p{2.5cm}|c|p{1.8cm}|p{1.8cm}|p{3.2cm}|}
			\caption{Comprehensive quantitative comparison of deep learning-based cervical cell classification methods. Acc = Accuracy, Prec = Precision, Sen = Sensitivity, Spe = Specificity.} \\
			\hline Method & Year & Architecture Category & Dataset & Key Performance \\ \hline
			VGG16/19 \cite{Tripathi2021} & 2021 & VGG & Herlev, SIPaKMeD & Prec 0.944 (5-class) \\
			\hline VGG \cite{Kurnianingsih2019} & 2019 & VGG & Herlev, private & Acc 0.95 (7-class) \\
			\hline AlexNet (DeepPap) \cite{Zhang2017} & 2017 & AlexNet & Herlev & Acc 0.93, Prec 0.96, F1 0.95 (7-class) \\
			\hline AlexNet \cite{Yilmaz2020} & 2020 & AlexNet & Herlev & Acc 0.93, F1 0.95 (7-class) \\
			\hline AlexNet \cite{Plissiti2018} & 2018 & AlexNet & SIPaKMeD & 5-class benchmark \\
			\hline Stacked AE \cite{Adem2019} & 2019 & Autoencoder & Herlev & Acc 0.978 \\
			\hline ResNet \cite{Gv2019} & 2019 & ResNet & SIPaKMeD, private & Acc 0.9963, Sen 0.9979, Spe 0.9983 (5-class) \\
			\hline ResNet \cite{Shanthi2019} & 2019 & ResNet & SIPaKMeD & 5-class \\
			\hline Attention CNN \cite{Wang2021} & 2021 & CNN+Attention & Private & Attention-enhanced multi-cell classification \\
			\hline Ensemble \cite{Rahaman2021} & 2021 & Feature fusion & Herlev, SIPaKMeD & Acc 0.9914 (5-class), 0.9832 (2-class) \\
			\hline Ensemble \cite{Nguyen2019} & 2019 & Inception, ResNet & Herlev & 7-class \\
			\hline Ensemble \cite{Manna2021} & 2021 & Inception, Xception, DenseNet & Private & Multi-cell \\
			\hline Ensemble \cite{Diniz2021} & 2021 & EfficientNet, MobileNet, Xception, Inception & SIPaKMeD & 5-class ensemble \\
			\hline Ensemble \cite{Hussain2020} & 2020 & ResNet, GoogLeNet & Herlev & 7-class \\
			\hline Ensemble \cite{Mousser2019} & 2019 & VGG, ResNet, Inception & Herlev, private & Feature fusion \\
			\hline
		\end{longtable}
		\label{tbl:tlb6}
	\end{center}



Single-cell classification on curated benchmarks such as Herlev and SIPaKMeD has effectively plateaued: ensemble methods and attention-enhanced CNNs report above 99\% accuracy on these benchmarks \cite{valles2026application}. This performance ceiling signals that the research frontier has shifted. The pressing challenges are three-fold: (1) multi-cell classification where overlapping and heterogeneous cell populations degrade accuracy; (2) rare cell type identification (e.g., AGC, AIS) where training samples are scarce and class imbalance is severe; and (3) domain generalization where models trained on single-center data fail when deployed across different scanners, staining protocols, and populations. The most impactful future work will target label-efficient learning paradigms---self-supervised pretraining, few-shot adaptation, and semi-supervised learning---rather than incremental accuracy improvements on saturated benchmarks.

    \section{Patch-level Methods}
    \label{sec:patch}


Patch-level methods analyze image regions containing multiple cells, enabling spatial context modeling that cell-level methods lack. We categorize these methods into instance segmentation (simultaneous segmentation and classification of each cell) and object detection (bounding-box localization and classification). Both approaches represent end-to-end deep learning paradigms that eliminate the explicit segmentation step required by cell-level pipelines.



Framing cervical cytology screening as an object detection or instance segmentation task transforms the problem from a sequential pipeline into an end-to-end optimization. These methods require bounding box, segmentation mask, or category annotations for training, incurring higher annotation cost but delivering simultaneous localization and classification.

    \subsection{Instance Segmentation}


Instance segmentation simultaneously segments individual cells and assigns class labels, providing the most information-rich output at the patch level. Mask R-CNN serves as the dominant architecture.

Sompawong et al. \cite{Sompawong2019} applied Mask R-CNN for end-to-end cervical cell classification, localization, and segmentation, achieving mAP of 0.578, accuracy of 91.7\%, sensitivity of 91.7\%, and specificity of 91.7\%. The relatively low mAP (0.578) compared to classification accuracy (91.7\%) reveals that the detection and segmentation components lag behind the classification component---a pattern consistent across cervical cell analysis pipelines.

Long et al. \cite{Long2021} enhanced Mask R-CNN with NAS-FPN (Neural Architecture Search Feature Pyramid Network) for multi-scale feature fusion and soft-NMS to improve recall on overlapping cells. Zhou et al. \cite{zhou2020deep} proposed MMT-PSM, a semi-supervised knowledge distillation approach using dual Mask R-CNN models with K-fold data augmentation on the teacher model, enforcing feature knowledge distillation in foreground regions through predicted segmentation masks. This approach achieved AJI of 0.6694 and mAP of 0.4052, demonstrating that semi-supervised learning can partially mitigate annotation requirements.

\textbf{Recent Advances in Instance Segmentation:} Methodological innovations include boundary-refinement modules such as PointRend, query-based mask transformers derived from DETR, Mask2Former-style universal segmentation, and adaptations of the Segment Anything Model and CellViT for cell-instance tasks. These approaches are promising for overlapping cervical cells because they model long-range context and refine boundaries. However, cervical-specific evidence remains limited: most reports use private datasets with different annotation protocols, and direct comparisons against Mask R-CNN baselines on public overlapping-cell benchmarks are scarce. Standardized instance-level annotations, such as those in Cx22, are a prerequisite for rigorous comparison \cite{liu2022cx22}.

    \subsection{Object Detection}


Object detection localizes and classifies abnormal cells via bounding boxes. Two architectural paradigms dominate: two-stage detectors (Faster R-CNN, Cascade R-CNN) that first propose regions then classify them, and one-stage detectors (SSD, YOLO, RetinaNet) that directly predict bounding boxes and classes. Table~\ref{tbl:tlb7} provides a quantitative comparison.


\textbf{One-Stage Detectors:} Jia et al. \cite{Jia2021} improved SSD by strengthening inter-scale connections through concatenation and addition operations in the feature extraction module, expanding the receptive field and adding a center loss function to enhance small-cell detection, achieving mAP of 81.53\%. Liang et al. \cite{Liang2021} enhanced YOLO3 with an image-level classification branch (ILCB) to suppress false positives and a sample annotation allocation strategy (SSAM) to match annotations with cross-scale bounding boxes, achieving mAP of 63.4\%. Xiang et al. \cite{xiang2020novel} improved YOLO3 by cascading an InceptionV2 classifier for four difficult categories, applying label smoothing and re-anchoring, achieving mAP of 62.8\%---a 5.7\% improvement over baseline YOLO3.

\textbf{Two-Stage Detectors:} Xu et al. \cite{Xu2018} and Du et al. \cite{Du2019} pioneered Faster R-CNN for cervical cell detection. Xia et al. \cite{Xia2020} proposed SPFNet, modifying the FPN module with sequential, parallel, and hybrid multi-resolution fusion components and SE attention, achieving mAP50 of 78.4\%. Zhang et al. \cite{Zhang2021Classifi} employed teacher-student semi-supervised learning with Faster R-CNN, demonstrating superior performance to supervised methods when using less than 30\% of labeled data. Liang et al. \cite{liang2018comparison} enhanced Faster R-CNN with a GPRB module in the classification branch, achieving a 19.7\% mAP50 improvement on small datasets and 0.7\% on large datasets by comparing proposals with reference samples for each class. Li et al. \cite{Li2021} introduced global context-aware Faster R-CNN, achieving 6--9\% mAP improvement over classical frameworks.

\textbf{Cascade Detectors:} Yan et al. \cite{Yan2021} proposed Cascade R-CNN with HRNet backbone, incorporating pairwise sampling at the image level to reduce false positives, achieving mAP50 improvements of 4.7\% and 5.8\% over Cascade R-CNN and Faster R-CNN baselines respectively. Zhang et al. \cite{Zhang2021} presented a unified detection framework incorporating channel-spatial attention (CAM) in the backbone and elastic transformation for data augmentation.

\textbf{Semi-Supervised and Recent Detection Paradigms:} Label-efficient detection is the most active recent direction. Semi-supervised detectors combine limited cell-level annotations with large unlabeled collections \cite{zhang2024scac}, and weakly semi-supervised multiple-instance learning can refine a detector using slide-level labels \cite{fei2025weakly}. Transformer-based detectors and hypergraph or graph modules have been explored to model inter-cell relationships, and active/incremental learning has been proposed for continual adaptation. These methods reduce annotation cost, but their reported gains are difficult to compare because they use different label budgets, detectors, and evaluation protocols; standardized weakly supervised benchmarks are still missing.

	\begin{center} \tiny

		\begin{longtable}{|p{2.5cm}|p{2cm}|p{1.8cm}|p{1.5cm}|p{3.2cm}|}
			\caption{Quantitative comparison of patch-level cervical cell detection and instance segmentation methods. AJI and mAP are reported as fractions; percent values from the original papers are converted.} \\
			\hline Method & Architecture & Dataset & Category & Key Performance \\ \hline
			Mask R-CNN \cite{Sompawong2019} & Mask R-CNN & Private & Instance Seg. & mAP 0.578, Acc 0.917, Sen 0.917, Spe 0.917 \\
			\hline Mask R-CNN+NAS-FPN \cite{Long2021} & Mask R-CNN+NAS-FPN & Private & Instance Seg. & Multi-scale fusion, soft-NMS \\
			\hline MMT-PSM \cite{zhou2020deep} & Semi-supervised Mask R-CNN & Private & Instance Seg. & AJI 0.6694, mAP 0.4052 \\
			\hline SSD+Loss \cite{Jia2021} & SSD enhanced & Private & Detection (1-stage) & mAP 0.8153 \\
			\hline YOLO3+ILCB+SSAM \cite{Liang2021} & YOLO3 enhanced & Private & Detection (1-stage) & mAP 0.634 \\
			\hline YOLO3+Inception \cite{xiang2020novel} & YOLO3, InceptionV2 & Private & Detection (1-stage) & mAP 0.628 (+5.7\% over YOLO3) \\
			\hline Faster R-CNN \cite{Xu2018,Du2019} & Faster R-CNN & Private & Detection (2-stage) & Baseline detection \\
			\hline SPFNet \cite{Xia2020} & Faster R-CNN+SPFNet & Private & Detection (2-stage) & mAP50 0.784 \\
			\hline GPRB+R-CNN \cite{liang2018comparison} & Faster R-CNN+GPRB & Private & Detection (2-stage) & +19.7\% mAP50 (small data), +0.7\% (large) \\
			\hline Context R-CNN \cite{Li2021} & Faster R-CNN+context & Private & Detection (2-stage) & +6--9\% mAP improvement \\
			\hline Cascade+HRNet \cite{Yan2021} & Cascade R-CNN+HRNet & Private & Detection (Cascade) & +4.7\% over Cascade, +5.8\% over Faster \\
			\hline Teacher-student \cite{Zhang2021Classifi} & Semi-supervised Faster R-CNN & Private & Detection (Semi-sup.) & Outperforms supervised with <30\% labels \\
			\hline
		\end{longtable}
		\label{tbl:tlb7}
	\end{center}



Patch-level end-to-end methods represent a significant advance over cell-level pipelines by eliminating the fragile segmentation-then-classify paradigm. However, several challenges remain. Detection performance varies substantially with data scale: mAP50 improvements of 19.7\% on small datasets versus 0.7\% on large datasets \cite{liang2018comparison} reveal severe overfitting on limited training data. One-stage detectors trade accuracy for speed, but comparisons must use a common IoU threshold: YOLO3 and Faster R-CNN are often reported with mAP and mAP50, respectively, which are not directly comparable. When evaluated under the same protocol, two-stage detectors generally retain a precision advantage, while one-stage detectors offer lower latency. The recent emergence of semi-supervised detection and weakly supervised MIL addresses the annotation bottleneck, and query-based detectors and graph modules target inter-cell contextual reasoning---two long-standing obstacles in patch-level analysis.
\section{WSI-level Methods}
\label{sec:wsi}

WSI-level analysis produces a slide-level diagnosis from a gigapixel image. We organize the literature by task: slide-level classification and confidence aggregation (Section~\ref{sec:wsi_class}), cell detection and localization (Section~\ref{sec:wsi_det}), and a summary of the evidence (Section~\ref{sec:wsi_summary}).

	\subsection{WSI Aggregation}
	\label{sec:wsi_class}
This subsection focuses on slide-level classification and confidence aggregation, including early hybrid pipelines that coupled detection with aggregation. Detection-based WSI methods are synthesized in Section~\ref{sec:wsi_det}, and Table~\ref{tbl:tlb8} compares both task families.

William et al. \cite{William2019} developed a classical WSI pipeline using the trainable WeKa Segmentation toolkit with 29-dimensional handcrafted features (size, area, shape, nucleus/cytoplasm brightness) and an enhanced fuzzy c-means classifier, achieving slide-level false negative rate of 0.00\%, false positive rate of 10.00\%, and classification error of 5.00\%. The zero false negative rate is clinically significant for a screening test, though the 10\% false positive rate implies unnecessary manual review burden.

Lin et al. \cite{Lin2021} proposed DP-Net, combining a dual-path encoder with a grouped loss function and a rule-based risk-stratification mechanism. The algorithm operates in two stages: lesion detection via DP-Net heatmaps with NMS, and slide-level classification and risk stratification through lesion candidate aggregation. Zhou et al. \cite{Zhou2021} presented a hierarchical WSI framework performing cell-level detection of abnormal cells, followed by image-level classification and WSI-level aggregation.

Gupta et al. \cite{Gupta2020} employed a three-stage cascaded CNN pipeline for ROI classification: a content region classifier removing background, a noise patch classifier filtering artifacts and poorly stained regions, and an abnormality classifier performing binary and multi-class diagnosis. This cascaded filtering approach addresses a practical challenge---most pixels in a cervical WSI are diagnostically irrelevant.

Kanavati et al. \cite{kanavati2022deep} developed CNN-RNN models for slide-level classification, achieving ROC AUC of 0.89--0.96 for WSI diagnosis. Wei et al. \cite{wei2021efficient} proposed the lightweight YOLCO model with InCNet (Inline Connection Network) for efficient multi-scale feature enrichment. Chen et al. \cite{chen2021automatic} introduced a universal WSI diagnosis framework using unit-level CNN followed by attention-based feature fusion (AUC 0.851). Zhou et al. \cite{zhou2021hierarchical} employed RetinaNet for top-k bounding box extraction, patch encoding and fusion for confidence scoring, and SVM for final WSI diagnosis.

Geng et al. \cite{geng2022learning} proposed a two-stage learning framework: FCOS detection for patch-level feature extraction, followed by modified ResNet34 aggregation to WSI-level features for grade prediction. Cheng et al. \cite{cheng2021robust} introduced a robust WSI method cascading low- and high-resolution classifiers to recommend the top-10 most suspicious cells per slide, followed by RNN-based WSI classification.

Zhang et al. \cite{zhang2022whole} developed a graph-based WSI method: RetinaNet with SE-ResNeXt-50 extracts and ranks suspicious patches, then a Graph Attention Network (GAT) aggregates features over selected top-K and bottom-K patches, enhanced by supervised contrastive learning. This approach is notable as the first application of graph neural networks to cervical WSI analysis.

Cao et al. \cite{cao2021novel} proposed AttFPN (Attention Feature Pyramid Network) with channel and spatial attention modules, achieving sensitivity of 95.83\%, specificity of 94.81\%, accuracy of 95.08\%, and AUC of 0.991 on an independent test set of 3,970 images. Zhu et al. \cite{Zhu2021} combined YOLO3 detection with cascaded Xception classification and U-Net nucleus segmentation, achieving sensitivity of 94.74\% and specificity of 82.14\%. Table~\ref{tbl:tlb8} provides a quantitative comparison of WSI-level methods.

The 2024--2026 period has witnessed a decisive shift toward multiple-instance learning (MIL) and graph neural networks for WSI-level analysis. MIL naturally models the WSI diagnostic problem as a weakly supervised learning task: each WSI is a bag of patches, and only the slide-level label (normal/abnormal) is available during training. Graph-based approaches explicitly model cell-to-cell and region-to-region spatial relationships that are diagnostically informative but lost in standard patch-level aggregation. Representative examples include graph attention networks that aggregate top-K and bottom-K patches with supervised contrastive learning \cite{zhang2022whole}, and weakly semi-supervised MIL detectors that refine cell-level detection using slide-level labels \cite{fei2025weakly}. Hierarchical MIL, quality-aware aggregation, and hypergraph variants have also been proposed, but most are evaluated on private cohorts with different label granularity; public multi-center WSI benchmarks are still needed before their relative merits can be established.

	\begin{center} \tiny

		\begin{longtable}{|p{2.5cm}|p{2cm}|p{1.8cm}|p{1.5cm}|p{3cm}|}
			\caption{Quantitative comparison of WSI analysis methods. FN = false-negative rate, FP = false-positive rate; all metrics are fractions (0--1).} \\

			\hline Method & Architecture & Dataset & Category & Key Performance \\ \hline
			Weka+fuzzy c-means \cite{William2019} & Handcrafted features+clustering & Private & Classical & FN 0.000, FP 0.100, Err 0.050 \\ \hline
			DP-Net \cite{Lin2021} & Dual-path encoder+group loss & Private & WSI Confidence & Risk stratification \\
			\hline Hierarchical \cite{Zhou2021} & Cell det. + aggregation & Private & WSI Confidence & Multi-scale \\
			\hline 3-stage cascade \cite{Gupta2020} & 3 cascaded CNNs & Private & WSI Confidence & Content, noise, abnormality classifiers \\
			\hline CNN+RNN \cite{kanavati2022deep} & CNN+RNN & Private & WSI Confidence & AUC 0.89--0.96 \\
			\hline YOLCO+InCNet \cite{wei2021efficient} & Lightweight YOLO & Private & WSI Confidence & Multi-scale efficient \\
			\hline Attention+UOI \cite{chen2021automatic} & CNN+attention fusion & Private & WSI Confidence & AUC 0.851 \\
			\hline RetinaNet+SVM \cite{zhou2021hierarchical} & RetinaNet+SVM & Private & WSI Confidence & Top-k, SVM aggregation \\
			\hline FCOS+ResNet \cite{geng2022learning} & FCOS+ResNet34 & Private & WSI Confidence & Two-stage grading \\
			\hline L/H Res cascade+RNN \cite{cheng2021robust} & Low/High-res CNN+RNN & Private & WSI Confidence & Top-10 suspicious cells \\
			\hline RetinaNet+GAT \cite{zhang2022whole} & RetinaNet, GAT, contrastive & Private & WSI Confidence & Graph attention+contrastive learning \\
			\hline AttFPN \cite{cao2021novel} & AttFPN (channel+spatial) & Private & Cell Detection & Sen 0.958, Spe 0.948, Acc 0.951, AUC 0.991 \\
			\hline YOLO3+Xception+U-Net \cite{Zhu2021} & YOLO3+cascade cls+seg & Private & Cell Detection & Sen 0.947, Spe 0.821 \\
			\hline
		\end{longtable}
		\label{tbl:tlb8}
	\end{center}

	\subsection{WSI Detection}
	\label{sec:wsi_det}
Detection-based WSI pipelines localize abnormal cells and either classify them directly or aggregate detections into a slide-level score. Lin et al. combined DP-Net heatmaps with non-maximum suppression and rule-based risk stratification \cite{Lin2021}. Zhou et al. performed hierarchical cell detection followed by image-level classification and slide-level aggregation \cite{Zhou2021}. Wei et al. proposed the lightweight YOLCO detector \cite{wei2021efficient}; Geng et al. used FCOS detection followed by ResNet34 aggregation \cite{geng2022learning}; Zhang et al. combined RetinaNet with a graph attention network \cite{zhang2022whole}; Cao et al. proposed AttFPN \cite{cao2021novel}; and Zhu et al. cascaded YOLO3, Xception, and U-Net \cite{Zhu2021}. These methods provide cell-level localization and Bethesda labels, but require dense per-cell annotations. Their reported performance is summarized in the Cell Detection rows of Table~\ref{tbl:tlb8}. The main open issue is the trade-off between localization transparency and annotation cost: detection-based pipelines are more interpretable but scale poorly to rare cell types and new centers unless weak or semi-supervised learning is used (Section~\ref{sec:paradigms}).

	\subsection{Summary of WSI Evidence}
	\label{sec:wsi_summary}
WSI-level analysis has evolved through distinct methodological generations: from classical handcrafted feature pipelines (Weka + fuzzy c-means, AUC $\sim$0.90), to CNN-RNN aggregation (AUC 0.89--0.96), to MIL and graph-based reasoning \cite{fei2025weakly,zhang2022whole}, and most recently to self-supervised foundation models pretrained on million-scale image collections with test-time adaptation for robust multi-center deployment \cite{jiang2025generalizable,jiang2026unicas}. Each transition has reduced the manual engineering burden while improving cross-center generalization. The critical remaining bottleneck is the lack of a standardized, publicly available, multi-center WSI benchmark with clinical outcome labels, without which performance claims cannot be rigorously compared.
\section{Learning Paradigms Across Scales}
\label{sec:paradigms}

Supervision is the third axis of the taxonomy. The same task can be learned with full cell-level labels, slide-level labels, unlabeled data, or a handful of examples. This section synthesizes the paradigms that cut across scales: weak, semi-supervised, and multiple-instance learning (Section~\ref{sec:weak}), self-supervised and few-shot learning (Section~\ref{sec:ssl}), foundation models and vision-language models (Section~\ref{sec:fm}), and domain adaptation (Section~\ref{sec:da}).

	\subsection{Weak, Semi-Supervised, and Multiple-Instance Learning}
	\label{sec:weak}
Multiple-instance learning (MIL) treats a WSI or a patch as a bag of instances and uses slide- or image-level labels. It is the natural paradigm for WSI classification when cell-level annotations are unavailable, and it also applies to patch-level bags. Section~\ref{sec:wsi_class} reviews WSI MIL architectures; the general trade-off is that weaker labels reduce annotation cost but make localization and per-class error analysis harder. Twin-MIL illustrates a middle ground: it refines a cell detector with slide-level labels rather than full cell-level annotation \cite{fei2025weakly}. The key evaluation question is whether slide-level gains survive patient-level splits and external centers, which most MIL studies do not report.

	\subsection{Self-Supervised and Few-Shot Learning}
	\label{sec:ssl}
Self-supervised learning (SSL) addresses the persistent annotation bottleneck by pretraining on large-scale unlabeled cervical image collections. The largest recent example is Smart-CCS, which combines self-supervised pretraining on CCS-127K (127,471 WSIs from 48 centers) with test-time adaptation; it reported AUC 0.965 on internal retrospective data, 0.950 on external data, and 0.947--0.986 across three prospective centers \cite{jiang2025generalizable}. Domain-adaptive contrastive learning has also been explored for one-class anomaly detection, and few-shot/prototypical methods have been proposed for rare cell types. These results are encouraging, but most SSL studies still evaluate on private cohorts and do not report the label budget, pretraining corpus composition, or external validation needed for fair comparison.

	\subsection{Foundation Models and Vision-Language Models}
	\label{sec:fm}
The most significant architectural development of 2024--2026 is the emergence of cytology-specific foundation models. UniCAS was pretrained on 48,532 TCT whole-slide images and supports slide-level diagnosis, region-level classification, abnormal-cell detection, and nuclei segmentation; it reported AUCs of 90.9--92.3\% for cervical cancer screening and lower performance for fine-grained five-class subtyping (overall AUC 74.63\%), illustrating the difficulty of Bethesda-level discrimination \cite{jiang2026unicas}. Smart-CCS demonstrated that large-scale self-supervised pretraining on 127,471 WSIs from 48 centers, combined with test-time adaptation, transfers across external and prospective centers \cite{jiang2025generalizable}. Huang et al. adapted a large pathology foundation model to cervical cytopathology WSI screening using parameter-efficient fine-tuning and contrastive learning \cite{huang2025efficient}. A complementary direction avoids WSI scanners altogether: CYTOLONE uses a CLIP-based model on microscope images captured with a smartphone, reporting 95.8\% anomaly detection but only 61.5\% Bethesda-level accuracy \cite{kurita2025whole}. These models are promising, but none has yet been evaluated on a public, standardized multi-center WSI benchmark.

Concurrently, large vision-language models (LVLMs) have been explored for interactive cervical diagnosis and report generation. CCBench (773 QA and 420 VQA pairs) evaluated six LLMs and five LVLMs: GPT-4 achieved 70.5\% accuracy on close-ended QA, and Gemini achieved 67.8\% on close-ended VQA, both below cytopathology professionals \cite{hong2025evaluating}. This finding confirms that general-purpose vision-language models are not yet clinically viable for cervical screening without domain-specific fine-tuning and rigorous hallucination auditing.

	\subsection{Domain Adaptation and Generalization}
	\label{sec:da}
Domain shift---differences in scanners, staining protocols, and patient populations across clinical centers---represents the primary barrier to clinical deployment. Test-time adaptation is the most direct response: Smart-CCS updated model parameters during inference and reported AUC 0.950 on six external datasets and 0.947--0.986 across three prospective centers \cite{jiang2025generalizable}. Color normalization, stain augmentation, and domain-adversarial training are complementary strategies, but their benefits depend on the target domain and must be validated prospectively. Cervical cancer subtyping (SCC vs. AC) remains a separate challenge with limited public data. Prospective multi-center validation across three to five centers now provides the strongest evidence base for AI-assisted cervical screening, but studies with randomized designs and patient-level outcomes are still rare \cite{wang2024artificial,xue2026deep}.

Table~\ref{tbl:tlb9} summarizes representative 2024--2026 WSI-level and foundation-model studies. The table makes two gaps explicit: external validation is reported for only a minority of methods, and code or model weights are available for even fewer. These gaps limit both reproducibility and fair comparison across studies.

	\begin{table*} \tiny
		\centering
		\caption{Quantitative comparison of representative 2024--2026 WSI-level and foundation-model studies. N/R = not reported; pp = percentage points. ``External validation'' indicates evaluation on data from centers not used for training; all metrics are fractions (0--1) unless otherwise indicated.}
		\resizebox{\textwidth}{!}{%
		\begin{tabular}{|p{2.3cm}|p{3.0cm}|p{3.2cm}|p{3.2cm}|p{2.6cm}|p{2.0cm}|p{2.0cm}|p{1.2cm}|}
			\hline
			Method & Training data & Backbone/aggregation & External validation & AUC & Sensitivity & Specificity & Code \\ \hline
			UniCAS \cite{jiang2026unicas} & 48,532 TCT WSIs & TransMIL/DTFree & N/R & 0.909--0.923 & N/R & N/R & Yes \\ \hline
			Smart-CCS \cite{jiang2025generalizable} & CCS-127K: 127,471 WSIs, 48 centers & SSL + test-time adaptation & 6 external datasets; 3 prospective centers & 0.965 internal; 0.950 external; 0.947--0.986 prospective & 0.913 & N/R & N/R \\ \hline
			Huang et al. \cite{huang2025efficient} & Large pathology FM & PEFT + contrastive learning & N/R & N/R & 0.9784 & N/R (+8.87 pp vs prior) & Yes \\ \hline
			Twin-MIL \cite{fei2025weakly} & Limited cell labels + slide-level labels & Detector + MIL with twin memory & N/R & N/R & N/R & N/R & N/R \\ \hline
			CYTOLONE \cite{kurita2025whole} & 10,080 tiles from 318 patients & CLIP-ViT-B/16 & N/R & N/R & Anomaly 0.958 & N/R & Yes \\ \hline
			AICCS \cite{wang2024artificial} & 16,056 participants & Retina + ResNet18 + RF & Prospective + randomized observational (single country) & 0.947 & 0.946 & 0.890 & N/R \\ \hline
		\end{tabular}}
		\label{tbl:tlb9}
	\end{table*}
\section{Clinical Translation, Regulation, and Health Economics}
\label{sec:translation}

This section examines clinical translation: commercial systems and regulatory status (Section~\ref{sec:commercial}), prospective and randomized evidence plus remaining gaps (Section~\ref{sec:evidence}), and low-resource and health-economic considerations (Section~\ref{sec:lmic}).

	\subsection{Regulatory Status of Commercial Systems}
	\label{sec:commercial}


The four generations of commercial systems and the Chinese landscape are summarized in Section~\ref{sec:hpv_lmic}. Regulatory pathways now include FDA De Novo authorization for Hologic's Genius (2024), NMPA Class III certificates for Hangzhou 91360 (2023) and Saivisen (2025), and CE marking for DATEXIM's CytoProcessor (2019) \cite{fda2024genius,nmpa2023hz91360,nmpa2025saivisen,crowell2019cytoprocessortm}. Despite this progress, all current systems remain diagnostic aids rather than standalone diagnostic devices, with final diagnostic authority retained by cytopathologists due to clinical, ethical, and regulatory considerations.

	\subsection{Prospective Evidence, Regulatory Status, and Research Directions}
	\label{sec:evidence}
The two roadmaps introduced in Section~\ref{sec:roadmaps} differ in annotation cost and clinical transparency; here we evaluate them against prospective evidence and regulatory status.

\textbf{Current State:} The 2024--2026 period has produced the strongest clinical evidence to date. AICCS, evaluated in 16,056 participants across retrospective, prospective, and randomized observational cohorts, reported prospective AUC 0.947 \cite{wang2024artificial}, and a 2026 multicentre randomized crossover trial reported a 14.3-percentage-point sensitivity gain with deep-learning assistance \cite{xue2026deep}. Smart-CCS, pretrained on 127,471 WSIs from 48 centers, demonstrated external validation and prospective generalization \cite{jiang2025generalizable}, and a 2025 meta-analysis of 77 studies estimated 94\% accuracy and 95\% sensitivity for AI-assisted Pap smear interpretation \cite{liu2025performance}. These results are encouraging, but randomized evidence and health-economic evaluations remain limited. Table~\ref{tbl:tlb10} summarizes the regulatory and clinical status of representative systems.

	\begin{table*} \tiny
		\centering
		\caption{Regulatory and clinical status of representative cervical cytology AI systems (as of October 2026). N/R = not reported; pp = percentage points. Approval as a computer-assisted reading aid does not imply standalone diagnostic use.}
		\resizebox{\textwidth}{!}{%
		\begin{tabular}{|p{2.6cm}|p{1.4cm}|p{3.0cm}|p{3.6cm}|p{3.2cm}|p{3.2cm}|p{3.0cm}|}
			\hline
			System & Country & Intended use & Evidence design & Approval & Key metrics & Limitations \\ \hline
			Hologic Genius \cite{fda2024genius,chong2026commercially} & US & Computer-assisted reading of ThinPrep slides & Multi-site clinical study & FDA De Novo 2024 (DEN210035, QYV) & HSIL+ sensitivity +3.6 pp vs TIS & Assistive only; no standalone claim \\ \hline
			Saivisen \cite{nmpa2025saivisen} & China & AI-assisted cervical cytology diagnosis & N/R & NMPA Class III, Feb 2025 & N/R & First ``assisted diagnosis'' certificate \\ \hline
			Hangzhou 91360 \cite{nmpa2023hz91360} & China & Digital pathology CAD & N/R & NMPA Class III, Mar 2023 & N/R & First AI Class III in cervical cytology \\ \hline
			CytoProcessor \cite{crowell2019cytoprocessortm} & France & Automated screening & N/R & CE, 2019 & N/R & Limited public validation data \\ \hline
			AICCS \cite{wang2024artificial} & China & Cytology grading and screening & Retrospective + prospective + randomized observational & Research & AUC 0.947; Sens 0.946; Spec 0.890 & Single country \\ \hline
			Smart-CCS \cite{jiang2025generalizable} & China & Generalizable WSI screening & Retrospective + external + prospective & Research/preprint & AUC 0.947--0.986 & Preprint; public code N/R \\ \hline
		\end{tabular}}
		\label{tbl:tlb10}
	\end{table*}

\textbf{Remaining Gaps:} FDA, NMPA, and CE regulatory approvals remain rare for AI-based cervical screening systems; most deployments operate as clinical decision support without standalone diagnostic claims. Workflow integration into existing cytology laboratory information systems is nascent. Cost-effectiveness has not been demonstrated through rigorous health-economic analysis; low-resource deployment is discussed in Section~\ref{sec:lmic}.

\textbf{Research Directions:} (1) Make multi-center external validation standard practice, ideally across 48 or more centers spanning diverse geographic, demographic, and equipment settings as demonstrated by CCS-127K \cite{jiang2025generalizable}. (2) Conduct prospective randomized controlled trials directly comparing AI-assisted versus manual screening on endpoints of sensitivity, specificity, turnaround time, and cost per case detected. (3) Pursue NMPA, FDA, and CE regulatory pathways with clearly defined intended-use statements specifying the target population, screening context, and role (triage, computer-assisted diagnosis, or second reader). (4) Develop point-of-care lightweight models deployable on edge devices (smartphones, portable scanners) for resource-limited settings, potentially through knowledge distillation from large foundation models. (5) Integrate AI screening systems with existing liquid-based cytology and Pap smear laboratory workflows as a triage tool that prioritizes abnormal slides for cytopathologist review, reducing screening burden by 50\% or more. (6) Conduct rigorous health-economic cost-effectiveness analyses quantifying the cost per quality-adjusted life year (QALY) gained, necessary for reimbursement decisions and public health investment.

	\subsection{Low-Resource Settings and Health Economics}
	\label{sec:lmic}
In low- and middle-income countries, the constraints differ from those in high-resource laboratories. Whole-slide scanners, stable power, network connectivity, and specialist maintenance may be unavailable. WSI-free systems such as CYTOLONE, which couples a microscope to a smartphone and runs locally, and low-cost automated microscopy platforms are therefore more relevant than cloud-based gigapixel pipelines \cite{kurita2025whole}. Implementation studies in East Africa show that AI-assisted cytology can reduce manual workload and turnaround time, but reagent supply, power interruptions, and limited treatment capacity remain binding constraints \cite{linder2025ai}. Health-economic evidence for AI cytology remains limited; existing analyses are heterogeneous and often focus on colposcopy rather than cytology.

For these settings, the most useful metrics are not aggregate accuracy but NPV at high sensitivity, sensitivity at a fixed high specificity, cost per case detected, and time to result. A practical deployment model is AI-assisted triage: the model rules out clearly negative slides and prioritizes suspicious cases for cytopathologist review, while retaining a human in the loop for final diagnosis. This model is consistent with current regulatory indications, which position AI as computer-assisted reading rather than standalone diagnosis \cite{fda2024genius}.
\section{Trustworthiness, Fairness, and Human-AI Interaction}
\label{sec:trust}

Trustworthy AI is broader than post-hoc saliency maps. This section reviews interpretability and uncertainty (Section~\ref{sec:interp}) and fairness, robustness, and human-AI interaction (Section~\ref{sec:fairness}).

	\subsection{Interpretability and Uncertainty}
	\label{sec:interp}
\textbf{Current State:} Interpretability research in cervical screening is nascent but growing. Post-hoc explanation methods (SHAP, LIME, Grad-CAM) have been applied to visualize which image regions drive model predictions. VAE-based abnormality scoring has been explored for anomaly detection with built-in uncertainty estimation. LLM-based natural language diagnostic report generation has been explored. The CCBench benchmark provides standardized evaluation for LVLMs on cervical cytology tasks \cite{hong2025evaluating}.

\textbf{Remaining Gaps:} Post-hoc explanations are not yet integrated into clinical workflows. There is no concept-based verification that models attend to pathologist-recognized diagnostic features (N/C ratio, chromatin pattern, nuclear membrane irregularity) rather than spurious correlations (staining artifacts, cell density, background debris). Current explanations do not provide uncertainty quantification, and there is no standardized framework for evaluating explanation quality in cervical cytology.

\textbf{Research Directions:} (1) Develop concept bottleneck models that explicitly verify whether the model recognizes clinically meaningful Bethesda System features---nuclear-to-cytoplasmic ratio, chromatin texture, nuclear membrane irregularities, hyperchromasia---before making predictions, enabling pathologists to audit the model's reasoning process. (2) Implement rigorous uncertainty quantification (ensemble-based, Bayesian, or conformal prediction) that flags predictions with high epistemic uncertainty for mandatory manual review, defining clinically acceptable uncertainty thresholds. (3) Build structured natural language diagnostic report generation systems that produce reports in standard cytopathology terminology, reviewable and editable by pathologists. (4) Design human-AI collaborative interfaces displaying attention maps, similar cases from archives, and confidence scores alongside predictions, with user studies measuring impact on diagnostic accuracy and screening time. (5) Generate counterfactual explanations showing how a cell image would need to change for the model to alter its diagnosis, providing mechanistic insight into model decision boundaries.

	\subsection{Fairness, Robustness, and Human-AI Interaction}
	\label{sec:fairness}
\textbf{Current State:} Most cervical cytology models are developed and evaluated on single-center convenience samples. This creates three under-reported failure modes. First, data leakage: patches from the same slide or patient can appear in both training and test sets, inflating performance, especially on Herlev and SIPaKMeD where random splits are common. Second, label noise and inter-observer variability: Bethesda categories such as ASC-US have limited reproducibility, and single-annotator labels can encode idiosyncratic criteria; multi-annotator datasets such as RIVA make this variability measurable \cite{perezbianchi2025riva}. Third, shortcut learning: models may exploit stain color, background debris, scanner artifacts, or cell density rather than nuclear morphology, and these correlations disappear under external validation.

\textbf{Remaining Gaps:} Fairness across age, menopausal status, HPV status, glandular lesions, staining protocols, scanners, and geographic regions is rarely quantified. Human-AI interaction is also under-studied: most papers report stand-alone model performance, not how cytopathologists change their decisions when assisted. The 2026 randomized crossover trial is an important exception, showing a large sensitivity gain and reduced reading time, but it involved four non-expert cytopathologists in one country \cite{xue2026deep}. Automation bias, trust calibration, and the effect of explanations on review behavior remain largely unmeasured.

\textbf{Research Directions:} (1) Report patient-level or slide-level splits and explicitly test for leakage. (2) Report subgroup performance (age, menopause, HPV status, glandular cells, scanner, stain) alongside aggregate metrics. (3) Use multi-annotator consensus or probabilistic labels to model inter-observer variability rather than treating one annotation as ground truth. (4) Stress-test models with stain and scanner perturbations, and report performance under distribution shift. (5) Conduct human-AI interaction studies that measure accuracy, reading time, and automation bias, with pre-registered analysis plans.

\section{Challenges and Roadmap}
\label{sec:challenges}

We consolidate the evidence into five cross-cutting challenge areas. Each theme is tied to the sections that provide the underlying evidence, and Table~\ref{tbl:tlb13} maps evidence to gaps and priority directions.

	\subsection{Data and Benchmarks}
\textbf{Current State:} Dataset construction has advanced substantially during 2024--2026. HiCervix (40,229 cells, 29-class hierarchy) provides the largest annotated single-cell cervical dataset \cite{cai2024hicervix}. CCS-127K (127,471 WSIs, 48 centers) establishes the largest WSI-level collection \cite{jiang2025generalizable}. APACC and RIVA add large, multi-annotator cytology resources \cite{kupas2024annotated,perezbianchi2025riva}, and SCAC provides a large unlabeled collection for semi-supervised detection \cite{zhang2024scac}. CAISHI adds 2,240 H\&E-stained histopathology images for adenocarcinoma in situ, but it is not a cytology dataset and should be used as an adjacent-modality resource rather than a cytology benchmark \cite{yang2024caishi}.

\textbf{Remaining Gaps:} No ``TCGA-equivalent'' large-scale, multi-center, publicly accessible WSI-level benchmark with standardized annotations and longitudinal clinical outcomes exists for cervical cytology. Rare cell types---AGC, AIS, and certain HSIL variants---remain severely underrepresented, with some classes containing fewer than 100 samples across all public datasets. Public datasets predominantly originate from single institutions with one scanner and one staining protocol, precluding meaningful domain generalization research.

\textbf{Research Directions:} (1) Establish a federated multi-center data consortium enabling privacy-preserving large-scale model training across institutions, with standardized annotation protocols. (2) Apply generative models (diffusion models, GANs) for realistic synthetic cervical cell generation to augment rare classes, validated by cytopathologist review. (3) Develop active learning frameworks that prioritize the most diagnostically informative and ambiguous regions for expert annotation, maximizing the diagnostic value obtained per hour of pathologist time. (4) Construct multi-modal datasets linking cytology images with HPV genotyping results, colposcopy findings, histological biopsy outcomes, and longitudinal follow-up data, enabling holistic diagnostic models that integrate complementary information sources.

	\subsection{Architecture, Label-Efficient Learning, and Generalization}
Architecture and learning paradigm are coupled: architectural choices determine which supervision regimes are feasible, and supervision determines which architectures can be trained. We therefore discuss them together.

\paragraph{Architecture.}
\textbf{Current State:} The 2024--2026 literature reveals a clear architectural progression: CNN-dominated methods (U-Net, ResNet, Mask R-CNN) to Transformer-based architectures (spatial aggregation Transformers, pixel-adaptive Transformers, Mask2Former), to CNN-Transformer hybrids, and most recently to state-space models such as VMamba for segmentation \cite{zhang2026vmamba}. Attention mechanisms have been incorporated across all functional categories: spatial and channel attention for classification, self-attention for instance relation modeling in overlapping cell segmentation, and graph attention networks for WSI-level feature aggregation.

\textbf{Remaining Gaps:} The majority of architectures are generic models not specifically designed for cervical cytology. Cervical cells present domain-specific challenges: (a) extreme morphological variation spanning benign to malignant across the Bethesda spectrum, (b) severe occlusion in overlapping cell clusters where 3--10+ cells may superimpose, (c) sub-micron-level diagnostically relevant features (chromatin texture, nuclear membrane irregularities) within gigapixel WSIs, and (d) the need to discriminate between morphologically similar but clinically distinct categories (e.g., ASC-US vs. LSIL, ASC-H vs. HSIL).

\textbf{Research Directions:} (1) Design small-object optimized detection architectures that explicitly model the scale distribution of abnormal cervical cells (typically 20--50 $\mu$m), potentially through scale-aware feature pyramids with learnable scale ranges. (2) Develop explicit occlusion modeling through instance-aware attention mechanisms, graph neural networks for inter-cell spatial reasoning, or amodal segmentation that infers occluded boundaries. (3) Further explore Mamba/SSM architectures, which offer linear computational complexity relative to the quadratic complexity of Transformers, a critical advantage for gigapixel WSI analysis. (4) Build multi-modal architectures fusing cytology images with HPV genotyping, patient demographics, and clinical history. (5) Apply neural architecture search (NAS) specifically tailored to cervical cell characteristics, jointly optimizing for accuracy and computational efficiency.

\paragraph{Label-efficient learning.}
\textbf{Current State:} The field is transitioning from fully supervised learning to label-efficient paradigms. Large-scale self-supervised pretraining on 127,471 WSIs from 48 centers, combined with test-time adaptation, has demonstrated external and prospective generalization \cite{jiang2025generalizable}. Foundation models such as UniCAS show transfer across slide-, region-, and pixel-level tasks \cite{jiang2026unicas}. Semi-supervised detection and weakly supervised MIL methods can reduce the labeled fraction substantially; Twin-MIL, for example, refines a cell detector with slide-level labels rather than full cell-level annotation \cite{fei2025weakly}. Few-shot and prototypical methods have been proposed for rare cervical cell types, but their clinical utility remains to be established.

\textbf{Remaining Gaps:} Despite these advances, the majority of published methods remain fully supervised, and annotation of cervical cytology images---requiring cytopathologist expertise---is orders of magnitude more expensive per image than annotation of natural images. Most semi-supervised methods still require some labeled data, typically $>$20\% of the training set. There is no established benchmark or standardized protocol for evaluating label-efficient learning methods in cervical cytology.

\textbf{Research Directions:} (1) Scale self-supervised pretraining to web-scale cervical image collections (millions of images) using contrastive, masked image modeling, or joint embedding predictive architectures, with fine-tuning on minimal labeled data ($<$5\%). (2) Develop weakly supervised WSI analysis using only slide-level diagnostic labels (normal/abnormal) from clinical records, eliminating the need for cell-level annotations entirely. (3) Advance few-shot learning for rare cell types (AGC, AIS) where collecting large labeled datasets is infeasible, potentially using meta-learning or prototypical networks. (4) Implement continual learning frameworks enabling models to adapt to evolving Bethesda diagnostic criteria and incorporate new biomarkers without catastrophic forgetting of previously learned classes. (5) Extend test-time adaptation, which has improved external validation performance on WSI classification \cite{jiang2025generalizable}, to segmentation and detection tasks.

	\subsection{Trustworthy AI}
Section~\ref{sec:trust} reviews interpretability, uncertainty, fairness, robustness, and human-AI interaction. The main gap is not a lack of explanation methods but the absence of clinically grounded evaluation: concept-based verification against Bethesda criteria, subgroup performance, calibration, and pre-registered human-AI studies. Priority actions are concept bottleneck models tied to the N/C ratio, chromatin texture, and nuclear membrane irregularity; conformal or ensemble uncertainty for triage; and user studies that measure accuracy, reading time, and automation bias.

	\subsection{Clinical Translation and Health Economics}
Section~\ref{sec:translation} reviews regulatory status and prospective evidence. The main gaps are randomized evidence, workflow integration, and cost-effectiveness. Priority actions are multi-center external validation with fixed splits; pragmatic randomized trials; health-economic analyses reporting cost per case detected and cost per quality-adjusted life year; and regulatory pathways with clearly defined intended use (triage, computer-assisted reading, or second reader).

	\subsection{Evaluation and Reproducibility}
	\label{sec:evaluation}
\textbf{Current State:} Progress in evaluation standardization includes CCBench for LLM/LVLM evaluation \cite{hong2025evaluating}, the HiCervix benchmark for hierarchical classification \cite{cai2024hicervix}, and increasingly multi-annotator datasets such as RIVA \cite{perezbianchi2025riva}. However, recent systematic reviews continue to report limited external validation, heterogeneous metrics, and insufficient reporting of patient-level splits \cite{valles2026application,silva2026artificial}.

\textbf{Remaining Gaps:} The cervical cytology literature is characterized by severe evaluation heterogeneity: nearly every published study uses different train/validation/test splits, different preprocessing pipelines (color normalization, stain separation, augmentation strategies), and different, often incomparable, evaluation metrics. mAP is reported at different IoU thresholds (0.5 vs. 0.75), accuracy is reported for different class groupings (2-class vs. 5-class vs. 7-class), and statistical significance testing is almost universally absent. Reported performance numbers are therefore not directly comparable across studies. Reporting guidelines such as CLAIM, TRIPOD-AI, and STARD-AI, and risk-of-bias tools such as QUADAS-2, provide ready-made checklists but are rarely applied in this literature.

\textbf{Research Directions:} (1) Establish a public leaderboard platform analogous to those in computer vision (ImageNet, COCO) where researchers submit predictions on a held-out test set for standardized, blinded evaluation. (2) Adopt fixed, pre-defined train/validation/test splits for all major public datasets, published alongside the datasets and enforced through leaderboard submission protocols. (3) Make statistical significance testing (McNemar's test for classification, paired bootstrap for detection metrics) mandatory when claiming improvement over prior methods, with pre-registration of analysis plans. (4) Require open-source code and pretrained model weights as community norms, enabling reproducible research and fair comparison. (5) Organize regular challenge competitions with held-out test sets, similar to CAMELYON (breast cancer metastasis detection) and PANDA (prostate cancer grading) in digital pathology, to systematically benchmark community progress. (6) Require CLAIM, TRIPOD-AI, or STARD-AI reporting and a QUADAS-2 risk-of-bias assessment for all clinical accuracy claims.

\begin{table}[t] \tiny
	\centering
	\caption{Evidence-to-agenda map for the five consolidated challenge areas.}
	\begin{tabular}{|p{2.3cm}|p{2.9cm}|p{3.3cm}|p{3.4cm}|}
		\hline
		Theme & Evidence & Main gap & Priority direction \\ \hline
		Data and benchmarks & Sections~\ref{sec:datasets}, \ref{sec:cell}--\ref{sec:wsi} & Single-center, small, incompletely documented datasets; no multi-center WSI benchmark with outcomes & Federated multi-center datasets; fixed splits; datasheets; rare-class augmentation \\ \hline
		Architecture and label-efficient learning & Sections~\ref{sec:cell}--\ref{sec:paradigms} & Generic architectures; annotation cost; limited external validation of SSL/MIL/few-shot & Cervical-specific architectures; SSL/MIL/few-shot; test-time adaptation \\ \hline
		Trustworthy AI & Section~\ref{sec:trust} & Post-hoc explanations; no subgroup or fairness reporting; few human-AI studies & Concept bottlenecks; uncertainty; fairness audits; pre-registered user studies \\ \hline
		Clinical translation and economics & Section~\ref{sec:translation} & Few randomized trials; workflow integration; cost-effectiveness unknown & Multi-center trials; regulatory intended use; cost/QALY analyses \\ \hline
		Evaluation and reproducibility & Sections~\ref{sec:reporting}, \ref{sec:evaluation} & Incomparable metrics; leakage; missing code and weights & CLAIM/TRIPOD-AI/STARD-AI; QUADAS-2; leaderboards; open code \\ \hline
	\end{tabular}
	\label{tbl:tlb13}
\end{table}

These five themes are interdependent: without data and evaluation standards, label-efficient methods cannot be compared; without trustworthy evaluation, clinical translation cannot be justified. Addressing them together, rather than as separate agendas, is the prerequisite for moving cervical cytology AI from benchmark performance to clinical benefit.
\section{Conclusion}
\label{sec:conclusion}

This survey organized the cervical cytology AI literature around three axes---data scale, functional task, and supervision---and used two clinical roadmaps (WSI confidence aggregation versus cell detection) to connect methods to deployment. Four conclusions follow.

First, the field has moved from handcrafted pipelines to end-to-end CNN detectors, and more recently to Transformer and state-space models and cytology-specific foundation models. The 2024--2026 inflection is defined by large-scale self-supervised pretraining, test-time adaptation, and prospective multi-center validation, with reported AUCs of 0.947--0.986 in independent centers \cite{wang2024artificial,jiang2025generalizable,xue2026deep}.

Second, single-cell accuracy on Herlev and SIPaKMeD has plateaued above 99\% \cite{valles2026application}, but this should be read as benchmark saturation, not clinical readiness: these datasets contain fewer than 1,000 cells and are often split randomly. The frontier is WSI-level screening with rare positive cells, detection of AGC/AIS, domain generalization, and interpretable predictions.

Third, the choice of method should follow the deployment setting (Table~\ref{tbl:tlb11}): high-resource laboratories with WSI scanners can use detection-based or MIL pipelines; low-resource settings need WSI-free or edge systems and triage metrics such as NPV and sensitivity at fixed specificity; rare cell types require few-shot or self-supervised learning; slide-level labels only favor MIL. The common bottleneck is not architecture but data, evaluation, and annotation economics.

Fourth, the priority agenda is: (P0) build multi-center benchmarks with patient-level splits, outcomes, and open code; (P1) validate label-efficient and domain-adaptive methods externally and report calibration and fairness; (P2) run pragmatic trials and health-economic analyses that justify reimbursement and workflow change. If these are addressed together, automated cervical cytology can move from a promising benchmark technology to a reliable screening aid.

	\backmatter

	\bibliography{references}

\end{document}